\documentclass{article}

\usepackage{times}
\usepackage{setspace}

\usepackage[dvips]{graphicx} 

\begin{document}
\begin{titlepage}
\title{The Cyborg Astrobiologist: First Field Experience}
\author{
Patrick Charles McGuire\thanks{corresponding author: {\it Email:} mcguire@physik.uni-bielefeld.de {\it or} mcguire@inta.es {\it Telephone:} +34 91 520 6432 {\it Fax:} +34 91 520 1621}\\
{\it Robotics \& Transdisciplinary Laboratories}\\
{\it Centro de Astrobiolog\'ia (CAB)}\\
{\it Carretera de Torrej\'on a Ajalvir km 4.5}\\
{\it Torrej\'on de Ardoz, Madrid, Spain 28850}\\
\and
Jens Orm\"o\\
{\it Planetary Geology Laboratory, CAB}
\and
Enrique D\'iaz Mart\'inez\\
{\it (formerly at) Planetary Geology Laboratory, CAB}\\
{\it (currently at) Direcci\'on de Geolog\'ia y Geof\'isica}\\
{\it Instituto Geol\'ogico y Minero de Espa\~na}\\
{\it Calera 1, Tres Cantos, Madrid, Spain 28760} \\
\and
Jos\'e Antonio Rodr\'iguez Manfredi, Javier G\'omez Elvira\\
{\it Robotics Laboratory, CAB}\\
\and
Helge Ritter, Markus Oesker, J\"org Ontrup\\
{\it Neuroinformatics Group, Computer Science Department}\\
{\it Technische Fakult\"at}\\
{\it University of Bielefeld}\\
{\it P.O.-Box 10 01 31}\\
{\it Bielefeld, Germany 33501}\\
\date{\today}
\vspace{3.0cm}
}
\maketitle
\thispagestyle{empty}
\end{titlepage}
\pagenumbering{arabic}
\begin{abstract}
We present results from the first geological field tests of the `Cyborg Astrobiologist', which is a wearable computer and video camcorder system that we are using to test and train a computer-vision system towards having some of the autonomous decision-making capabilities of a field-geologist and field-astrobiologist. The Cyborg Astrobiologist platform has thus far been used for testing and development of these algorithms and systems: robotic acquisition of quasi-mosaics of images, real-time image segmentation, and real-time determination of interesting points in the image mosaics.  The hardware and software systems function reliably, and the computer-vision algorithms are adequate for the first field tests. In addition to the proof-of-concept aspect of these field tests, the main result of these field tests is the enumeration of those issues that we can improve in the future, including: first, detection and accounting for shadows caused by 3D jagged edges in the outcrop; second, reincorporation of more sophisticated texture-analysis algorithms into the system; third, creation of hardware and software capabilities to control the camera's zoom lens in an intelligent manner; and fourth, development of algorithms for interpretation of complex geological scenery. Nonetheless, despite these technical inadequacies, this Cyborg Astrobiologist system, consisting of a camera-equipped wearable-computer and its computer-vision algorithms, has demonstrated its ability of finding genuinely interesting points in real-time in the geological scenery, and then gathering more information about these interest points in an automated manner.
  \end{abstract}

\markright{The Cyborg Astrobiologist: First Field Experience: McGuire et al.}\
\thispagestyle{myheadings}
\newpage

\markright{The Cyborg Astrobiologist: First Field Experience: McGuire et al.}\

\begin{spacing}{1}
\section{Introduction}
\begin{spacing}{1}
{\it ``With great autonomy comes great responsibility''}

 --with apologies to Peter Parker's Uncle Ben (Lee {\it et al.}\ 2002, 2004).
\end{spacing}

\vspace{0.5cm}
Outside of the Mars robotics community, it is commonly presumed that
the robotic rovers on Mars are controlled in a time-delayed joystick manner, wherein commands are sent to the rovers several if not many times per day, as new information is acquired from the rovers' sensors.  However, inside the Mars robotics community, they have learned that this process is rather cumbersome, and they have developed much more elegant methods for robotic control of the rovers on Mars, with highly significant degrees of robotic autonomy. Particularly, the Mars Exploration Rover (MER) team has demonstrated autonomy for the two robotic rovers Spirit \& Opportunity to the level that: practically all commands for a given Martian day (1 `sol' $=$ 24.6 hours) are delivered to each rover from Earth before the robot wakens from its power-conserving nighttime resting mode (Arvidson {\it et al.}\ 2003; Maki {\it et al.}\ 2003; Bell {\it et al.}\ 2003; Squyres {\it et al.}\ 2003; Crisp {\it et al.}\ 2003). Each rover then follows the commanded sequence of moves for the entire sol, moving to desired locations, articulating its arm with its sensors to desired points in the workspace of the robot, and acquiring data from the cameras and chemical sensors. From an outsider's point of view, these capabilities may not seem to be significantly autonomous, in that all the commands are being sent from Earth, and the MER rovers are merely executing those commands. But upon closer inspection, what the MER team has achieved is truly amazing:
\begin{spacing}{1}
\begin{trivlist}
\raggedright
\item $\bullet$ Firstly, the rovers can move to points $50-150$ meters away in one sol with autonomous obstacle avoidance enabled for the uncertain or dangerous parts of the journey (Goldberg, Maimone \& Matthies 2002; Nesnas, Maimone \&  Das 1999).
\item $\bullet$ Secondly, prior to a given sol, based upon information received after the previous sol, the MER team has the remarkable capabilities to develop a command sequence of tens or hundreds of robotic commands for the entire sol. As of July 4, 2004, this was taking 4-5 hours per sol for the mission team to complete, rather  than the 17 hours per sol that it took at the beginning of the MER missions (Squyres, unpublished\footnote{lecture notes from the course on the MER rovers from Astrobiology Summer School {\it ``Planet Mars''}, Universidad Internacional Menendez Pelayo Santander, Spain}, 2004).
\end{trivlist}
\end{spacing}

Such capabilities for semi-autonomous teleoperated robotic `movement and discovery' are a significant leap beyond the capabilities of the previous Mars lander missions of Viking I \& II and of Pathfinder \& Sojourner, particularly in the large improvements of the mobility of both the cameras and the instrumentation onboard the robot, allowing significant discoveries to be made, and new insights to be obtained (Chan {\it et al.}\ 2004; Squyres {\it et al.}\ 2004; Squyres \& Athena Science Team 2004; Moore 2004; Catling 2004; Crumpler {\it et al.}\ 2004; Golembek {\it et al.}\ 2004; Squyres 2004; McSween {\it et al.}\ 2004; Christensen {\it et al.}\ 2004; Arvidson {\it et al.}\ 2004; Greeley {\it et al.}\ 2004; Herkenhoff {\it et al.}\ 2004; Bell {\it et al.}\ 2004).  Nonetheless, we would like to build upon this great success of the MER rovers by developing enhancing technology that could be deployed in future robotic and/or human exploration missions to the Moon, Mars, and Europa.

\markright{The Cyborg Astrobiologist: First Field Experience}

 One future mission deserves special discussion for the technology developments described in this paper: the Mars Science Laboratory, planned for launch in 2009 (MSL'2009). A particular capability desired for this MSL'2009 mission will be to rapidly traverse to up to three geologically-different scientific points-of-interest within the landing ellipse (Heninger {\it et al.}\footnote{Heninger, B., M. Sander, J. Simmonds, F. Palluconi, B. Muirhead, C. Whetsel, ``Mars Science Laboratory Mission 2009 Landed Science Payload'', Proposal Information Package (draft version),   http://centauri.larc.nasa.gov/msl/PIP-Drft\_FBO-RevA-031113.pdf}, unpublished, 2003).  These three geologically-different sites will be chosen from Earth by analysis of relevant remote-sensing imagery from a possible co-launched, satellite-based, Synthetic Aperture Radar (SAR) system or high-resolution maps from prior missions (i.e.\ 30 cm/pixel spatial resolution from the HiRISE imager on the Mars Reconnaisance Explorer with imagery expected in 2006-2007, McEwen {\it et al.}\ 2003). Possible desired maximal traversal rates could range from 300-2000 meters/sol in order to reach each of the three points-of-interest in the landing ellipse in minimum time.  These are our rough estimates of the typical maximum traversal rates, based upon current capabilities, mission goals, increased rover size and expectations of the state of technology for the possible 2009 mission;  our estimate is partially based upon such work as described in: Goldberg, Maimone \& Matthies (2002); Olson, Matthies, Schoppers \& Maimone (2003); Crawford (unpublished\footnote{``NASA Applications of Autonomy Technology'', NASA Ames Research Center, Intelligent Data Understanding Seminar,  http://is.arc.nasa.gov/IDU/slides/Crawford\_Aut02c.pdf}, 2002); and Crawford \& Tamppari (unpublished\footnote{``Mars Science Laboratory -- Autonomy Requirements Analysis'', NASA Ames Research Center, Intelligent Data Understanding Seminar,  http://is.arc.nasa.gov/IDU/slides/Crawford\_MSL02c.pdf}, 2002).

Given these substantial expected traversal rates of the MSL'2009 rover, autonomous obstacle avoidance (Goldberg, Maimone \& Matthies 2002) and autonomous visual odometry \& localization (Olson, Matthies, Schoppers \& Maimone 2003) will be essential to achieve these rates, since otherwise, rover damage and slow science-target approach would be the results.  Given such autonomy in the rapid traverses, it behooves us to enable the autonomous rover with sufficient scientific responsibility. Otherwise, the robotic rover exploration system might drive right past an important scientific target-of-opportunity along the way to the human-chosen scientific point-of-interest. Crawford \& Tamppari (previous footnote, 2002) and their team summarize possible `autonomous traverse science', in which every 20-30 meters during a 300 meter traverse (in their example), science pancam and Mini-TES (Thermal Emission Spectrometer) image mosaics are autonomously obtained. They state that ``there {\it may be} onboard analysis of the science data from the pancam and the mini-TES, which compares this data to predefined signatures of carbonates or other targets of interest. If detected, traverse may be halted and information relayed back to Earth.'' This onboard analysis of the science data is precisely the technology issue that we have been working towards solving. This paper is the first detailed published report describing our progress towards giving a robot some aspects of autonomous recognition of scientific targets-of-opportunity (see McGuire {\it et al.}\ (2004) for a brief description and motivation of our work).

 \begin{figure}[ht]
 \center{\includegraphics[height=5.0cm]{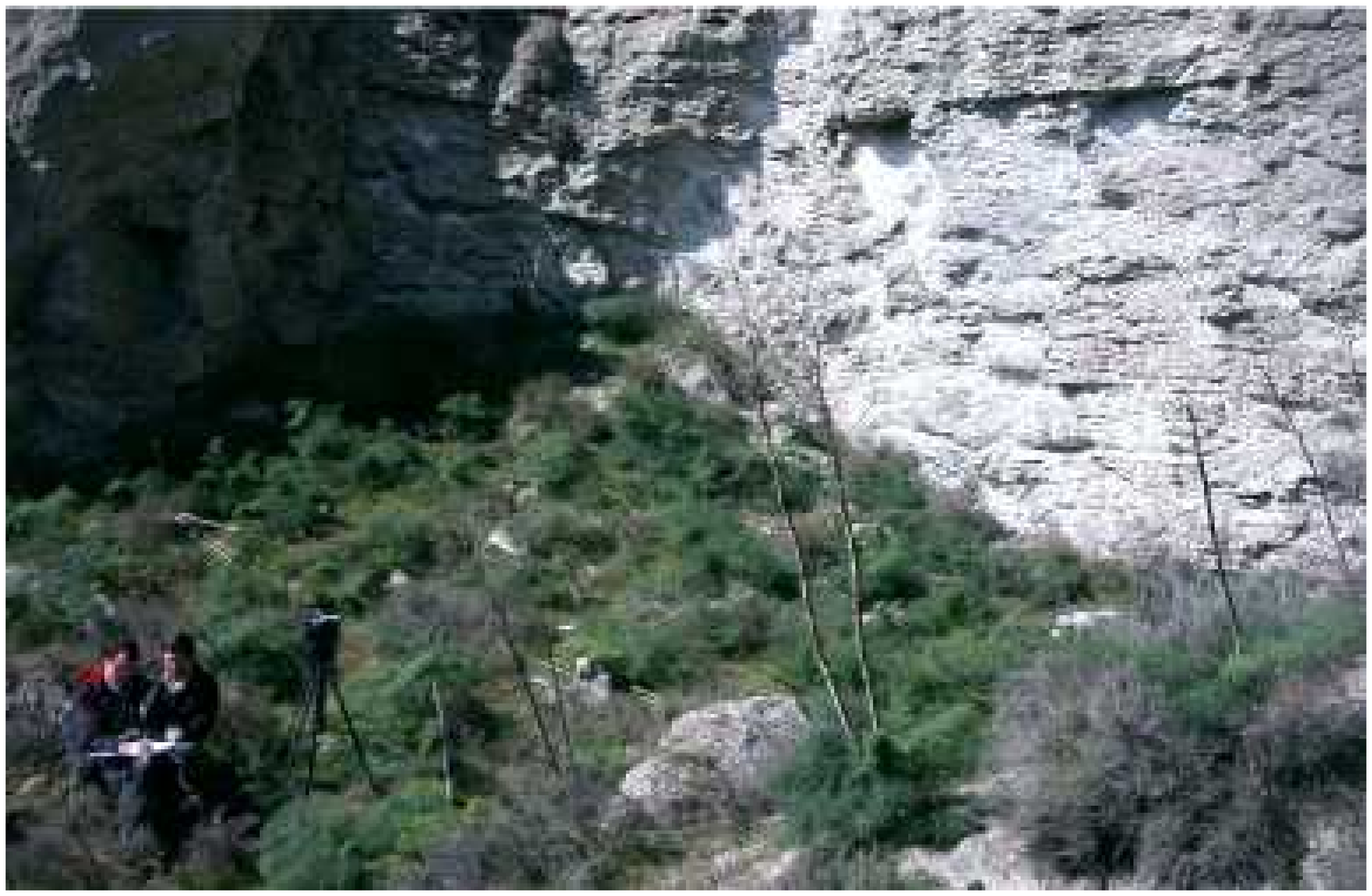}
 \includegraphics[height=5.0cm]{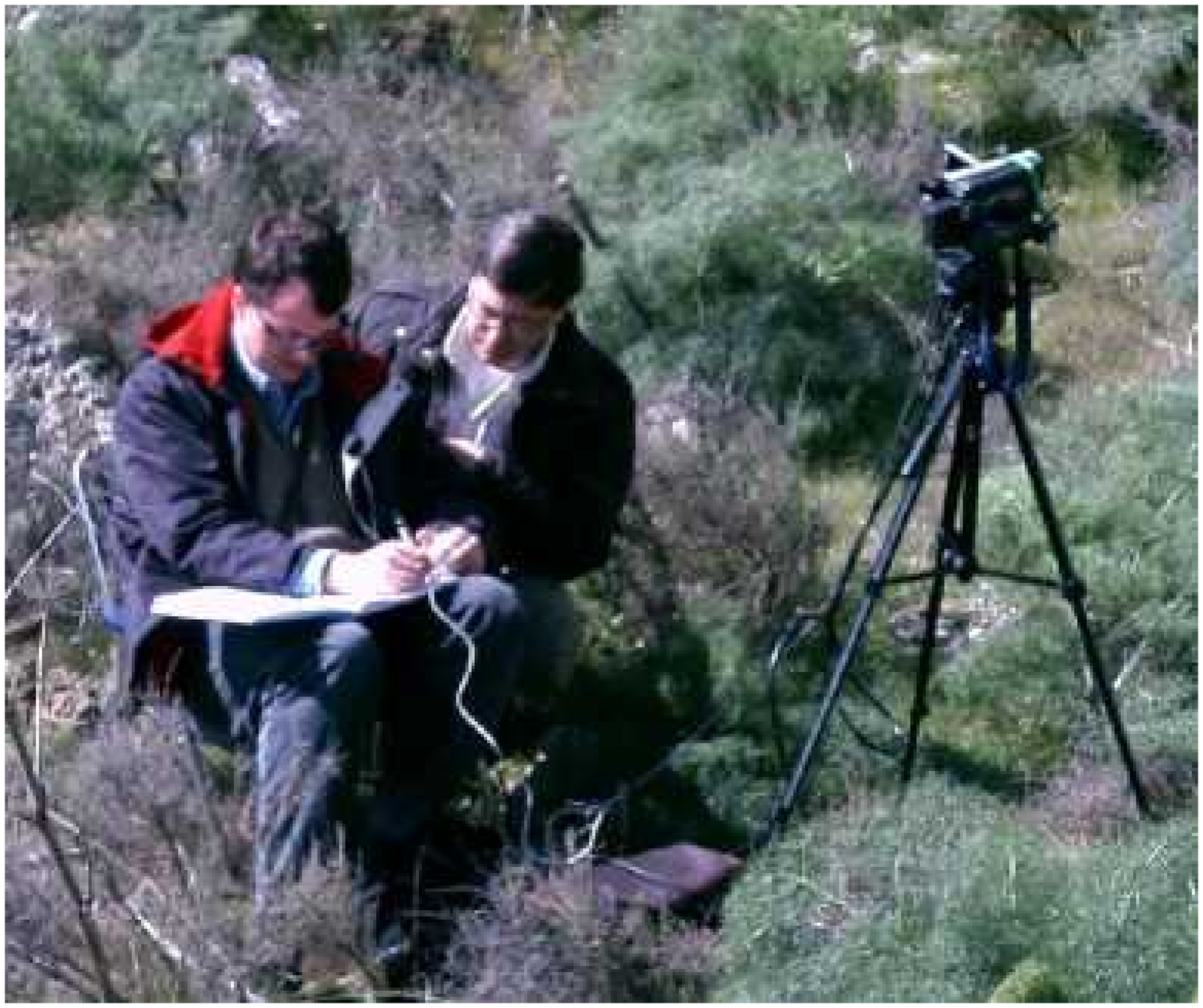}}
  \caption{ On the left, D\'iaz Mart\'inez \& McGuire with the Cyborg Astrobiologist
System at 1:30PM on 3 March 2004, 10 meters from the outcrop cliff
that is being studied during the first geological field mission near Rivas Vaciamadrid. We are taking notes
prior to acquiring one of our last-of-the-day mosaics and its set of interest-point chips. On the right is a close-up view of the same scene, but focusing on the half-seated/half-tripod Cyborg Astrobiologist and the assisting human Geologist. Photo copyright: D\'iaz Mart\'inez, Orm\"o \& McGuire}
 \end{figure}

Our general aim of study for this project is to develop autonomous computer vision algorithms for natural terrestrial geological scenery in uncontrolled lighting conditions (see our `Cyborg Astrobiologist' research platform in Figure 1), with the expectation that such algorithms will mature in the future so that they can be used in the human or robotic astrobiological exploration of Mars and some of the moons of our solar system.  This project is not only aimed at giving more scientific autonomy to robotic explorers of the surfaces of other planetary bodies, but it is also meant to take some of the burden off of the large team of human geologists at mission control here on the Earth by performing some of the low-level geological image analysis autonomously, thus freeing the human geologists to do more high-level analysis of the scenery. Such a system could also assist human geologists here in the field on the Earth for their studies, as well as astronauts in their geological field studies (who may be less-practiced in geology and more burdened with other duties and concerns). Our long-term aim of study is for the Cyborg Astrobiologist system to be able to understand the geological history of an outcrop in order to decide which area should be approached for sampling. Our short-term aim of study is to demonstrate that the Cyborg Astrobiologist system can (in the field) find the most unique regions in an image mosaic, and then to autonomously study those interesting regions in more detail.

 In this report, we give additional introductory background in Section 1.1, a discussion of the hardware and software for our ``Cyborg Astrobiologist'' system in Section 2, summaries \& results of the first two geological expeditions of our Cyborg Astrobiologist system in Sections 3 \& 4, followed by a more general discussion \& conclusions in Section 5.

\subsection{Autonomous Recognition of Scientific Astrobiological Targets-of-Opportunity}
 We discuss here two of the related efforts in the development of autonomous recognition of scientific targets-of-opportunity for astrobiological exploration: firstly, the work on developing a Nomad robot to search for meteorites in Antartica led by the Carnegie Mellon University Robotics Institute (CMU-RI) (Apostolopoulos {\it et al.}\ 2000; Pederson 2001; Wagner 2000; Vandapel {\it et al.}\ 2000), and secondly, the work by a group at NASA Ames Research Center (NASA-ARC) on developing a Geological Field Assistant (GFA) (Gulick {\it et al.}\ (2001, 2002a, 2002b, 2003, 2004)).  There are several other efforts in the community which also deserve mention (including Volpe 2003; Huntsberger {\it et al.}\ 2002; Storrie-Lombardi, Grigolini {\it et al.}\ 2002; Corsetti \& Storrie-Lombardi 2003; Cabrol {\it et al.}\ 2001; Wettergreen {\it et al.}\ 1999; Whittaker {\it et al.}\ 1997; Cheeseman {\it et al.}\ 1988; Cheeseman \& Stutz\ 1996). Outside of the astrobiology community, there is some similar work on automated obstacle avoidance (instead of target discovery) by Batavia \& Singh (2001), among others.  

 The CMU-RI team has programmed the capabilities to autonomously distinguish between rocks and ice/snow in various lighting conditions (direct sunlight, shadows, and diffuse light in overcast conditions) into the science computer of its meteorite-searching Nomad robot (Apostolopoulos {\it et al.}\ 2000). This autonomous decision between whether a pixel is ice/snow or if it is rock is decided by how `blue' the pixel is: if the pixel has its `blue' ratio ($blue/(blue + green)$) above a certain lighting-dependent threshold, then it is considered to be ice/snow; otherwise, it is considered to be rock/meteorite. This system works rather well except when the rocks are in shadows (but we naively suggest that a non-linear separator in the 2D space of `blue' and intensity could solve this problem even in shadows). After a rock/meteorite is segmented from the ice/snow based upon how `blue' it is, it can then trigger the Nomad robot's Bayesian classifier network system to compute the probability of the rock being any one of six different meteor types or of 19 different terrestrial rock types, based upon data from a metal detector, from a color camera detector, and from a spectrometer. The Nomad robot ``[autonomously] found and [autonomously] classified five indigenous meteorites [of 42 rocky targets studied and classified over a 2500 m$^2$ search area] during an expedition to a remote site of Elephant Moraine [in Antarctica] in January 2000.'' (Apostolopoulos {\it et al.}\ 2000).

 The NASA-ARC GFA team have developed a number of algorithms for assisting an astronaut in doing geology or in adding capabilities to an autonomous robotic geologist (Gulick {\it et al.}\ (2001, 2002a, 2002b, 2003, 2004)). They have been building an image and spectral database of rocks and minerals [now there are over 700 such rock and mineral entries in their database] for continued development of advanced classifier algorithms, including future development work to autonomously identify `key igneous, sedimentary and some metamorphic rocks' (Gulick {\it et al.}\ 2003). As of 2003, they were able to identify ``some igneous rocks from texture and color information; quartz, silica polymorphs, calcite, pyroxene and jarosite from both visible/near-infrared and mid-infrared spectra; and high-iron pyroxenes and iron-bearing minerals using visible/near-infrared spectra only'' (Gulick {\it et al.}\ 2003). As of 2004, the NASA Jet Propulsion Laboratory (JPL) CLARAty system team (Volpe 2004) had incorporated the NASA-ARC GFA team's rock, layer and horizon detectors (Gulick {\it et al.}\ 2001) for ``possible use by MSL'2009 and [other] future Mars robotics missions'' (Gulick {\it et al.}\ 2004). Work reported in 2004 includes significant progress on their igneous rock detection system: the Decision Tree method has yielded better and more decipherable results than the Bayesian Network approach, giving ``correct identification of greater than $80\%$ of granites and granodiorites and greater than $80\%$ of andesites and basalts, using color and texture algorithms combined [without spectral information]'' (Gulick {\it et al.}\ 2004).

\section{The Cyborg Geologist \& Astrobiologist System}
\markright{The Cyborg Astrobiologist: System}
 Our ongoing effort in this area of autonomous recognition of scientific targets-of-opportunity for field geology and field astrobiology is beginning to mature as well. To date, we have developed and field-tested a GFA-like ``Cyborg Astrobiologist'' system (McGuire {\it et al.}\ 2004) that now can:
\begin{spacing}{1}
\begin{trivlist}
\raggedright
\item $\bullet$ Use human mobility to maneuver to and within a geological site and to follow suggestions from the computer as to how to approach a geological outcrop;
\item $\bullet$ Use a portable robotic camera system to obtain a mosaic of color images;
\item $\bullet$ Use a `wearable' computer to search in real-time for the most uncommon regions of these mosaic images;
\item $\bullet$ Use the robotic camera system to re-point at several of the most uncommon areas of the mosaic images, in order to obtain much more detailed information about these `interesting' uncommon areas;
\item $\bullet$ Use human intelligence to choose between the wearable computer's different options for interesting areas in the panorama for closer approach; and
\item $\bullet$ Repeat the process as often as desired, sometimes retracing a step of geological approach. 
\end{trivlist}
\end{spacing}

   In the Mars Exploration Workshop in Madrid in November 2003, we demonstrated some of the early capabilities of our `Cyborg' Geologist/Astrobiologist System (McGuire {\it et al.}\ 2004). We have been using this Cyborg system as a platform to develop computer-vision algorithms for recognizing interesting geological and astrobiological features, and for testing these algorithms in the field here on the Earth.

   The half-human/half-machine `Cyborg' approach (see Figures 1 \& 5) uses human locomotion and human-geologist intuition/intelligence for taking the computer vision-algorithms to the field for teaching and testing, using a wearable computer. This is advantageous because we can therefore concentrate on developing the `scientific' aspects for autonomous discovery of features in computer imagery, as opposed to the more `engineering' aspects of using computer vision to guide the locomotion of a robot through treacherous terrain. This means the development of the scientific vision system for the robot is effectively decoupled from the development of the locomotion system for the robot.

   After the maturation of the computer-vision algorithms, we hope to
transplant these algorithms from the Cyborg computer to the on-board
computer of a semi-autonomous robot that will be bound for Mars or one of the interesting moons in our solar system.

\subsection{Geological Approach to an Outcrop; Metonymical Contacts} 
\markright{The Cyborg Astrobiologist: System: Geological Approach}
 This method of using a semi-autonomous system to guide a geological approach to an outcrop, in which the wearable computer decides or gives options as to what in the panorama is the most interesting and deserves closer study, is meant to partially emulate the decision process in the mind of a professional field geologist. An approach from a distance, using information available at further distances to guide the approach to closer distances, is a reasonable, and possibly, generic method used by most field geologists when approaching an outcrop subject to study. Our technique of presenting several point-like options for more close-up study is a reasonable method to `linearize' the convoluted thought processes of a practicing field geologist. 

 \begin{figure}[th]
 \center{\includegraphics[width=15cm]{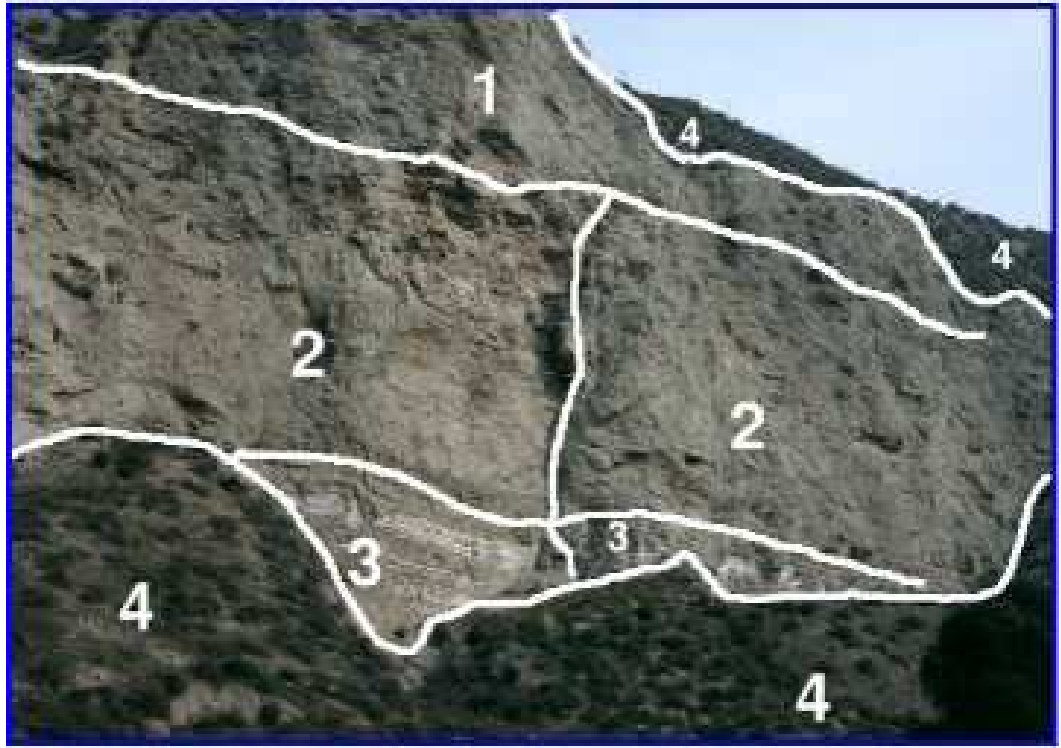}}
  \caption{An image segmentation made by geologist, D\'iaz Mart\'inez, of the outcrop during the first mission to Rivas Vaciamadrid. Region 1 has a tan color and a blocky texture; Region 2 is subdivided by a vertical fault and has more red color and a more layered texture than Region 1; Region 3 is dominated by white and tan layering; and Region 4 is covered by vegetation. The dark \& wet spots in Region 3 were only observed during the second mission, 3 months later. The Cyborg Geologist/Astrobiologist made its own image segmentations for portions of the cliff face that included the area of the white layering at the bottom of the cliff (see Figures 6-8).  Photo copyright: D\'iaz Mart\'inez, Orm\"o \& McGuire}
 \end{figure}

McGreevy (1992, 1994) has completed significant detailed cognitive \& ethnographic studies of the practice of field geology, using a head-mounted camera and virtual-reality visor system for part of his study as well as extensive on-site interviews of field-geologist colleagues. One particularly interesting concept from McGreevy's work is that of `metonymic' geological contacts. For field-geology, this concept of metonymy is the juxtaposition of differing geological units as well as the relation between them; metonymy is to differences, as metaphor is to similarities.

Both of the field geologists on our team have also independently stressed the importance to field geologists of such geological contacts and the differences between the geological units that are separated by the geological contact. For this reason, in March 2003, the roboticist and the field geologists decided that the most important tool to develop for the beginning of our computer vision algorithm development was that of  `image segmentation'. Such image segmentation algorithms would allow the computer to break down a panoramic image into different regions (see Figure 2 for an example), based upon similarity, and to find the boundaries or contacts between the different regions in the image, based upon difference. Much of the remainder of this paper discusses the first geological field trials with the wearable computer of the segmentation algorithm we have developed over the last year.

\subsection{Image Segmentation, Uncommon Maps, Interest Maps, and Interest Points}
\markright{The Cyborg Astrobiologist: System: Image Analysis}

With human vision, a geologist:
\begin{spacing}{1}
\begin{trivlist}
\raggedright
\item$\bullet$ Firstly, tends to pay attention to those areas of a scene which are most unlike the other areas of the scene; and then,
\item$\bullet$ Secondly, attempts to find the relation between the different areas of the scene, in order to understand the geological history of the outcrop.
\end{trivlist}
\end{spacing}

The first step in this prototypical thought process of a geologist was our motivation for inventing the concept of uncommon maps. See Figure 3 for a simple illustration of the concept of an uncommon map.
We have not yet attempted to solve the second step in this prototypical thought process of a geologist, but it is evident from the formulation of the second step, that
human geologists do not immediately ignore the common areas of the scene.  Instead, human geologists catalog the common areas and put them in the back of their minds for
``higher-level analysis of the scene'', or in other words, for determining explanations for the relations of the uncommon areas of the scene with the common areas of the scene.

 \begin{figure}[th]
 \center{\includegraphics[width=12cm]{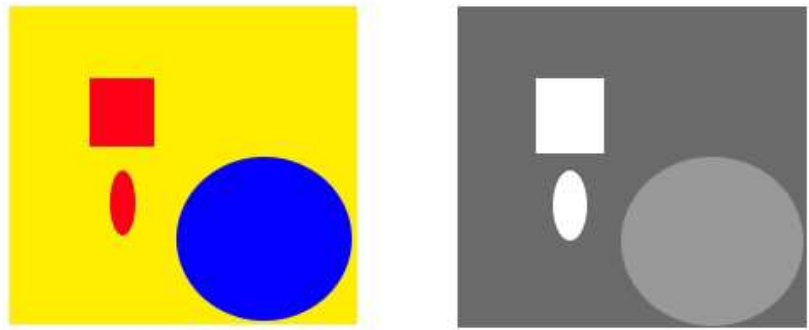}}
  \caption{For the simple, idealized image on the left, we show the corresponding uncommon map on the right. The whiter areas in the uncommon map are more uncommon than the darker areas in this map. }
 \end{figure}

 Prior to implementing the `uncommon map', the first step of the prototypical geologist's thought process, we needed a segmentation algorithm, in order to produce pixel-class maps to serve as input to the uncommon map algorithm.
We have implemented the classic co-occurrence histogram algorithm (Haralick, Shanmugan \& Dinstein 1973; Haddon \& Boyce 1990). For this work, we have not included texture information in the segmentation algorithm and the uncommon map algorithm. Currently, each of the three bands of $HSI$ color information is segmented separately (see Section 2.4 and the appendix), and later merged in the interest map by summing three independent uncommon maps. Maybe in future work, image segmentation simultaneously using color \& texture could be developed for and tested on the Cyborg Astrobiologist System (e.g., Freixenet, Mu\~noz, Mart\'i \& Llad\'o 2004).

The concept of an `uncommon map' is our invention, though it probably has been independently invented by other authors, since it is somewhat useful. In our implementation, the uncommon map algorithm takes the top 8 pixel classes determined by the image segmentation algorithm, and ranks each pixel class according to how many pixels there are in each class.  The pixels in the pixel class with the greatest number of pixel members are numerically labelled as `common', and the pixels in the pixel class with the least number of pixel members are numerically labelled as 'uncommon'.  The `uncommonness' hence ranges from 1 for a common pixel to 8 for an uncommon pixel, and we can therefore construct an uncommon map given any image segmentation map.  Rare pixels that belong to a pixel class of 9 or greater are usually noise pixels in our tests thus far, and are currently ignored. In our work, we construct several uncommon maps from the color image mosaic, and then we sum these uncommon maps together, in order to arrive at a final interest map. This summing to determine an interest map has been also studied by Rae, Fislage \& Ritter (1999). In our implementation for this paper, the interest map is a convenient name to call the sum of several uncommon maps, so the interest map is in a sense also an uncommon map. However, in general, the interest map could also sum other maps as well; and in the implementation of Rae, Fislage \& Ritter (1999), their `cortical' interest map algorithm not only sums several different types of maps, but the algorithm also adapts the weighting coefficients for these maps in order to highlight the summand map(s) which are most salient, most uncommon, or most interesting.  For our work, we have not activated this adaptive summing of the interest map algorithm,  and we only use uncommon maps in the interest map sum. 

In this paper, we develop and test a simple, high-level concept of interest points of an image, which is based upon finding the centroids of the smallest (most uncommon) regions of the image. Such a `global' high-level concept of interest points differs from the lower-level `local' concept of interest points (F\"orstner 1986; F\"orstner \& G\"ulch 1987) based upon corners and centers of circular features. However, this latter technique with local interest points is used by the MER team for their stereo-vision image matching and for their visual-odometry and visual-localization image matching (Arvidson {\it et al.}\ 2003; Goldberg, Maimone \& Matthies 2002; Olson, Matthies, Schoppers \& Maimone 2003; Nesnas, Maimone \& Das 1999). Our interest point method bears somewhat more relation to the higher-level wavelet-based salient points technique (Sebe, Tian, Loupias, Lew \& Huang 2003), in that they search first at coarse resolution for the image regions with the largest gradient, and then they use wavelets in order to zoom in towards the salient point within that region that has the highest gradient. Their salient point technique is edge-based, whereas our interest point is currently region-based. Since in the long-term, we have an interest in geological contacts, this edge-based \& wavelet-based salient point technique could be a reasonable future interest-point algorithm to incorporate into our Cyborg Astrobiologist system for testing.

 For a review of some of the general aspects of computer vision for natural scenery see: Batlle, Casals, Freixenet \& Mart\'i (2000).  We give a description of our specific software system in Section 2.4 and more details in the Appendix.

\begin{figure}[th]
\center{\includegraphics[width=10cm]{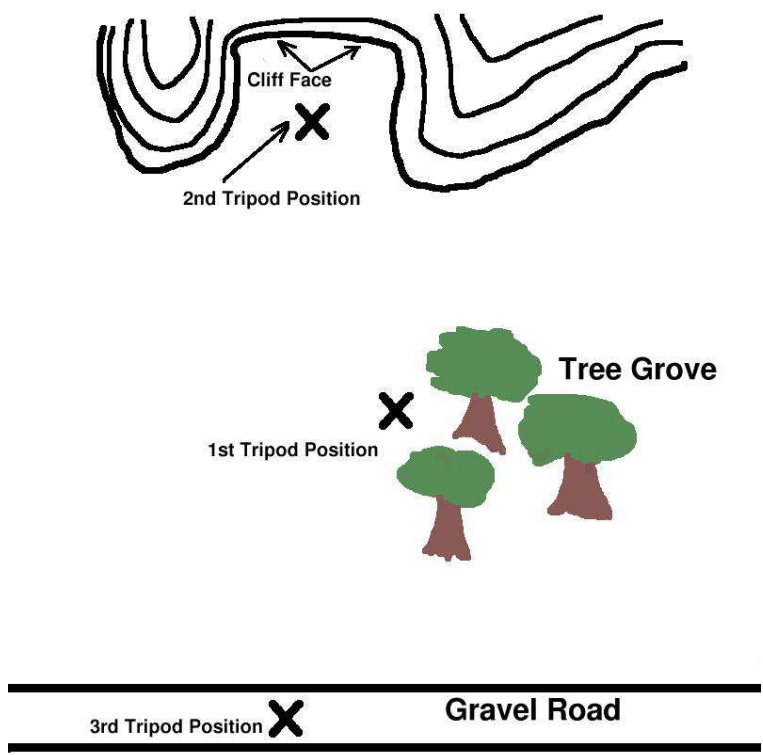}}
 \caption{ Map of the three tripod positions of the Cyborg Astrobiologist system during the second mission to Rivas Vaciamadrid (this map is not to scale). See Section 3.3 for a description of the geological approach suggested by the Cyborg Astrobiologist to the outcrop. See Figure 1 for an image of the system in action during the first mission, when it was near the 2nd tripod position shown here. See Figure 5 for an image of the system in action during the first mission when it was near the 1st tripod position shown here. Figures 6, 7 \& 8 show mosaics obtained during the second mission for the 3rd, 1st \& 2nd (respectively) tripod positions shown here; these same Figures (6, 7 \& 8) also show the higher resolution color imagery of the top three interest points for each of the image mosaics.
}
\end{figure}

\subsection{Hardware for the Cyborg Astrobiologist}
\markright{The Cyborg Astrobiologist: System: Hardware}

The non-human hardware of the Cyborg Astrobiologist system consists of:
\begin{spacing}{1}
\begin{trivlist}
\raggedright
\item $\bullet$ a 667 MHz wearable computer (from ViA Computer Systems in Minnesota) with a `power-saving' Transmeta `Crusoe' CPU and 112 MB of physical memory, 
\item $\bullet$ an {\it SV-6} Head Mounted Display (from Tekgear in Virginia, via the Spanish supplier Decom in Val\`encia) with native pixel dimensions of 640 by 480 that works well in bright sunlight,
\item $\bullet$ a SONY `Handycam' color video camera (model {\it DCR-TRV620E-PAL}),
\item $\bullet$ a thumb-operated USB finger trackball from 3G Green Green Globe Co., resupplied by ViA Computer Systems and by Decom,
\item $\bullet$ a small keyboard attached to the human's arm,
\item $\bullet$ a tripod for the camera, and
\item $\bullet$ a Pan-Tilt Unit (model {\it PTU-46-70W}) from Directed Perception in California with a bag of associated power and signal converters.
\end{trivlist}
\end{spacing}

 The power-saving aspect of the wearable computer's Crusoe processor is important because it extends battery life, meaning that the human does not need to carry very many spare batteries, meaning less fatigue for the human geologist. A single lithium-ion battery, which weighs about 1kg, can last for 3 hours or so for this application.  The SONY Handycam provides real-time imagery to the wearable computer via an IEEE1394/Firewire communication cable. The current system uses five independent batteries, one for each sub-system: the computer, the head-mounted display, the camera, the Pan-Tilt Unit, and the communication interface between the Pan-Tilt Unit and the computer. Assuming fully-charged batteries, all batteries are currently more than sufficient for a mission of several hours.

 The main reason for using the tripod is that it allows us to use the Pan-Tilt Unit for mosaicking and for re-pointing. Without the tripod, there would be an unusual amount of jitter in the image-registration caused by the human fidgeting with the camera, regardless of whether the camera is hand-mounted, shoulder-mounted or head-mounted. In future studies, if the mosaicking or re-pointing is not desired, then neither the Pan-Tilt Unit nor the tripod would be necessary; eliminating both the tripod and the Pan-Tilt Unit would certainly add to the mobility of the Cyborg Astrobiologist system.

 For this particular study, which included the Pan-Tilt Unit and the tripod, the wearable computer could have been replaced by a suitable non-wearable computer. However, we would like to retain flexibility in our work for future studies, in which the mobility granted by the wearable computer may be essential. The head-mounted display was much brighter than our earlier tablet display, and together with the thumb-operated finger mouse, the head-mounted display was more ergonomic than than our earlier tablet display and stylus. However, the spatial resolution of the head-mounted display was somewhat less than the resolution of the tablet display, but sufficient for the current task. It would be useful to be able to switch rapidly between the tablet display and the head-mounted display, but our system is not capable of this at this moment.

 The thumb-operated finger trackball worked rather well for navigating the Microsoft Windows environment of the wearable system. However, when used outdoors in bright sunlight, we often had to shade the trackball with the user's second hand in order for it to function properly. Since this mouse is an optical mouse, the mouse fails when sufficient sunlight leaks into the trackball. Alternatively, we could have sealed the seams in the plastic mouse with tape or glue. Additionally, since the wearable computer only has one USB port, the system would have been easier to use when downloading data to a USB memory stick if the USB finger trackball had been the PS2 model of the finger trackball; this will be changed before the next mission.

 Two other issues affected the usage \& mobility of the wearable system in the field. First, the serial port of the ViA wearable computer often failed to provide a signal of sufficient amplitude; this signal is needed to drive the Pan-Tilt Unit used for robotic exploration of the environment. We sent the wearable computer system back to the manufacturer for investigation, and they suggested that we reinstall the COM port in software when we had this problem. But since this problem was so frequent (especially after putting the wearable computer into sleep mode for power conservation), we decided to fix the signal problem with an external Operational Amplifier (OpAmp). This external OpAmp system functions well, but it requires occasional 9V battery changes, and it needs a certain amount of engineering improvements in ergonomics and miniaturization. Second, the Pan-Tilt Unit requires +12 to +30V of power. The battery we have chosen for this task weighs about 2kg, and lasts for 5-10 hours with the currently-used Cyborg software. The battery, together with the signal conversion box (from Serial signals to PTU signals) and the signal OpAmp box, require an extra, small carrying bag in order to use the PTU in the field. This bag is usually placed on the ground when each tripod position is reached, but for transport between the tripod positions the PTU bag is added baggage and puts some minor limits on mobility. The PTU system certainly merits further engineering.

\subsection{Software for the Cyborg Astrobiologist}
\markright{The Cyborg Astrobiologist: System: Software}

 The wearable computer processes the images acquired by the color digital video camera, to compute a map of interesting areas. The computations include: simple mosaicking by image-butting, as well as two-dimensional histogramming for image segmentation (Haralick, Shanmugan \& Dinstein 1973; Haddon \& Boyce 1990).  This image segmentation is independently computed for each of the Hue, Saturation, and Intensity (H,S,I) image planes, resulting in three different image-segmentation maps. These image-segmentation maps were used to compute `uncommon' maps (one for each of the three (H,S,I) image-segmentation maps): 
each of the three resulting uncommon maps gives highest weight to those regions of smallest area for the respective (H,S,I) image planes. Finally, the
three (H,S,I) uncommon maps are added together into an interest map, which is used by the Cyborg system for subsequent interest-guided pointing of the camera. The image-processing algorithms and robotic systems-control algorithms are all programmed using the graphical programming language, NEO/NST (Ritter {\it et al.}\footnote{Ritter, H., {\it et al.}\ (1992, 2002).  ``The Graphical Simulation Tookit, Neo/NST'', http://www.TechFak.Uni-Bielefeld.DE/techfak/ags/ni/}, unpublished, 1992, 2002). Using such a graphical programming language adds certain flexibility and ease-of-understanding to our Cyborg Astrobiologist project, which is by its nature largely a software project. We discuss some of the details of the software implementation in the appendix.

  After segmenting the mosaic image (see the appendix and Figure 9), it becomes obvious that a very simple method to find interesting regions in an image is to look for those regions in the image that have a significant number of uncommon pixels. We accomplish this by (Figure 10 for an example): first, creating an uncommon map based upon a linear reversal of the segment peak ranking (we have also studied nonlinear reversals of the segment ranking); second, adding the 3 uncommon maps (for $H$, $S$, \& $I$) together to form an interest map; and third, blurring this interest map\footnote{with a gaussian smoothing kernel of width $B=10$. This smoothing kernel effectively gives more weight to clusters of uncommon pixels, rather than to isolated, rare pixels.}. 

Our study extends Rae, Fislage \& Ritter's interest map concept by using uncommon maps based upon image segmentation of $H$, $S$ \& $I$ for the component maps of the interest map, rather than using the $H$, $S$ \& $I$ maps themselves for the component maps of the interest map. This extension is especially important for our purposes of finding the most uncommon and interesting areas of an image. By using such an image-fusion technique as Rae, Fislage \& Ritter's Cortical Interest Map technique to combine three uncommon/segmentation maps we also avoided the complexity of developing a true image-segmentation algorithm for {\it color} images in order to determine a single uncommon/segmentation map. If the current work is sufficiently valued by the Astrobiology or other communities, then maybe the Cyborg platform could be used for the development, testing, and optimization of true color image-segmentation algorithms.

  Based upon the three largest peaks in the blurred/smoothed interest map, the Cyborg system then guides the Pan-Tilt Unit to point the camera at each of these three positions to acquire high-resolution color images of the three interest points, as shown in four different examples in Figures 6-8. The robotic control and image mosaicking portion of the Cyborg software required significant time for development and debugging since the logic of this autonomy is somewhat complicated. However, by extending a simple image-acquisition and image-processing system to include robotic and mosaicking elements, we were able to conclusively demonstrate that the system can make reasonable decisions by itself in the field for robotic pointing of the camera (Figures 6-8).

\section{Descriptive Summaries of the Field Site and of the Expeditions}
\markright{The Cyborg Astrobiologist: Field Site and Expeditions}

   On the 3rd of March, 2004, three of the authors (McGuire, D\'iaz Mart\'inez \& Orm\"o) tested the ``Cyborg Astrobiologist" system for the first time at a geological site, the gypsum-bearing southward-facing stratified cliffs near the ``El Campillo" lake of Madrid's Southeast Regional Park (Castilla Ca\~namero\footnote{Castilla Ca\~namero, G. (2001). Informe sobre las pr\'acticas profesionales realizadas en el Centro de Educaci\'on Ambiental: El Campillo. Internal report to the Consejer\'ia de Medio Ambiente de la Comunidad de Madrid.}, unpublished, 2001; Calvo, Alonso \& Garcia del Cura 1989; IGME 1975), outside the suburb of Rivas Vaciamadrid.

 After post-mission analysis of the results from the Cyborg Astrobiologist's first mission, a second mission was planned for the 12th of May, 2004. After problems were discovered in preparation for the second mission, we decided to delay the second mission to Rivas Vaciamadrid until the 11th of June, 2004. For the second, rescheduled mission on the 11th of June, McGuire \& Orm\"o took the Cyborg Astrobiologist system to the same southward-facing stratified cliffs outside of Rivas. Due to the significant storms in the previous 3 months, there were 2 points in the gypsum cliffs that were leaking water down the face of the cliff (despite the hot weather before \& during the second mission). 

\subsection{Description and Choice of the Field Site at Rivas Vaciamadrid}
The rocks at the outcrop are of Miocene age (15-23.5 Myrs before present), and consist of gypsum and clay minerals, also with other minor minerals present (anhydrite, calcite, dolomite, etc.). These rocks form the whole cliff face which we studied, and they were all deposited in and around lakes with high evaporation (evaporitic lakes) during the Miocene. They belong to the so-called Lower Unit of the Miocene.  Above the cliff face is a younger unit (which we did not study) with sepiolite, chert, marls, etc., forming the white relief seen above from the distance, which is part of the so-called Middle Unit or Intermediate Unit of the Miocene.  The color of the gypsum is normally grayish, and in-hand specimens of the gypsum contain large and clearly visible crystals.

We chose this locality for several reasons, including:
\begin{spacing}{1}
\begin{trivlist}
\raggedright
\item $\bullet$ It is a so-called `outcrop', meaning that it is exposed bedrock (without vegetation or soil cover);
\item $\bullet$ It has distinct layering and significant textures, which will be useful for this study and for future studies;
\item $\bullet$ It has some degree of color differences across the outcrop;
\item $\bullet$ It has tectonic offsets, which will be useful for future high-level analyses of searching for textures that are discontinuously offset from other parts of the same texture;
\item $\bullet$ It is relatively close to our workplace;
\item $\bullet$ It is not a part of a noisy and possibly dangerous construction site or highway site.
\end{trivlist}
\end{spacing}

The upper areas of the chosen portion of the outcrop at Rivas Vaciamadrid (see Figure 2) were mostly of a tan color and a blocky texture. There was some faulting in the more layered middle areas of the outcrop, accompanied by some slight red coloring. The lower areas of the outcrop were dominated by white and tan layering. Due to some differential erosion between the layers of the rocks, there was some shadowing caused by the relief between the layers.  However, this shadowing was perhaps at its minimum for direct lighting conditions, since both expeditions were taken at mid-day and since the outcrop runs from East to West. Therefore, by performing our field-study at Rivas during mid-day, we avoided possible longer shadows from the sun that might occur at dawn or dusk.

   D\'iaz Mart\'inez had studied similar geological sites in the region of Rivas Vaciamadrid, prior to the missions. He explains the bright white area at the bottom of the cliffs as being due to mineral efflorescences (most probably calcium and magnesium sulfates) on the {\it surface} of the cliff rock. This cliff rock is mostly gypsum and clays. The efflorescence is just a surface feature formed by evaporation of the water seeping out from the rock pores by capillarity.

\subsection{Details from the First Mission to Rivas}
\markright{The Cyborg Astrobiologist: Field Site and Expeditions: First Mission}

   We arrived at the site at 10:30am on March 3, and for the next hour, the two geologists of our team both made independent expert-geologist assessments of the area.  In their assessments, these two geologists included their reasoning processes of how they would gradually approach and study the site, from an initial distance of about 300 meters down to a final distance of physical contact with the geological structures. An analysis of the geologists' assessments will be made available in the near future. However, in Figure 2, we show the segmentation of the outcrop, according to the geologist D\'iaz Martinez, for reference and as a sample of the geologists' assessments.

   At about 11:30am on the same day, the roboticist finished setting up the Cyborg Astrobiologist system, and started taking mosaic images of the cliff face, with the computer-controlled \& motorized color video camera. It generally takes some time to set up the system (in this case one hour) because:
\begin{spacing}{1}
\begin{trivlist}
\raggedright
\item $\bullet$ Several cables need to be attached and secured,
\item $\bullet$ The computer needs to boot,
\item $\bullet$ Three programs need to be activated and checked,
\item $\bullet$ Communication between the serial port and the Pan-Tilt Unit needs to be confirmed, and finally
\item $\bullet$ The non-computer-controlled zoom factor on the camera-lens needs to be set and optimized for the geological scenery, and the Pan-Tilt Unit's step size needs to be set to match the camera's zoom factor. 
\end{trivlist}
\end{spacing}

 The computer was worn on the user's belt, and typically took 3-5 minutes to acquire and compose a mosaic image composed of $M \times N$ subimages. Typical values of $M \times N$ used were $3 \times 9$ and $11 \times 4$. The sub-images were downsampled in both directions by a factor of 4-8 during these tests; the original sub-image dimensions were $360 \times 288$.

   Several mosaics were acquired of the cliff face from a distance of about 300 meters, and the computer automatically determined the three most interesting points in each mosaic. Then, the wearable computer automatically repointed the camera towards each of the three interest points, in order to acquire {\it non-downsampled} color images of the region around each interest point in the image. All the original mosaics, all the derived mosaics and all the interest-point subimages were then saved to hard disk for post-mission study. During the entire expedition, the robotics engineer could observe what the wearable computer system was doing through a very bright, sunlight-readable head-mounted computer display (HMD). The robotics engineer could \& did make adjustments on-the-fly to the automated-exploration settings and algorithms by using this HMD in conjunction with a hand-held free-movement thumbmouse and a small arm-mounted keyboard.

 Three other tripod positions were chosen for acquiring mosaics and interest-point image-chip sets. At each of the three tripod positions, 2-3 mosaic images and interest-point image-chip sets were acquired. One of the chosen tripod locations was about 60 meters (see Figure 5) from the cliff's face; the other two were both about 10 meters (see Figure 1) from the cliff face. There was some concern that if we approached much closer for any significant amount of time, then rocks from the fragile gypsum-bearing cliffs could fall upon us (there was plenty of evidence of past rock falls). We kept a distance, but noticed no such rock falls during the time we were there.

\subsection{Details from the Second Mission to Rivas}
\markright{The Cyborg Astrobiologist: Field Site and Expeditions: Second Mission}

   Orm\"o \& McGuire arrived at 12pm on June 11. One of us first suspected that somebody had lit some fires near the bottom of the cliffs, since there were two black stains, which were medium-sized and soot-like, covering in total about 10-15\% (about 4-6 m$^2$) of the bright white area at the bottom of the cliffs.  Closer analysis discounted this suspicion, as the black stains were really caused by two separate leaks of water\footnote{audible from 60 meters distance.} from the cliff face, with each leak wetting \& darkening a small region of the cliff face.

  Due to the hot sun, we chose our first Cyborg Astrobiologist tripod position (see the map in Figure 4) to be in the cool shade of the grove of trees about 60 meters from the cliff face (See Figure 5). We stayed at this tripod position for about 1.3 hours, acquiring several $2\times2$ mosaics, and several $4\times3$ mosaics, with image-downsampling of 4. The system most often determined the wet spots (Figure 7) to be the most interesting regions on the cliff face. This was encouraging to us, because we also found these wet spots to be the most interesting regions\footnote{These dark \& wet regions were interesting to us partly because they give information about the development of the outcrop. Even if the relatively small spots were only dark, and not wet (i.e., dark dolerite blocks, or a brecciated basalt), their uniqueness in the otherwise white \& tan outcrop would have drawn our immediate attention. Additionally, even if this had been our first trip to the site, and if the dark spots had been present during this first trip, these dark regions would have captured our attention for the same reasons. The fact that these dark spots had appeared after our first trip and before the eecond trip was not of paramount importance to grab our interest (but the `sudden' appearance of the dark spots between the two missions did arouse our higher-order curiosity).}.


  To end the mission, we chose two other tripod positions (see the map in Figure 4) for the Cyborg Astrobiologist to acquire mosaics, to search for interest points and then to record detailed information about the interest points. The first of these two tripod positions was about 10 meters from the cliff face, which was significantly closer than the original tripod position in the grove of trees and was approximately the same position as in the first mission, as depicted in Figure 1.  During this `close-up' study of the cliff face, we intended to focus the Cyborg Astrobiologist exploration system upon the two points that it found most interesting when it was in the more distant tree grove, namely the two wet and dark regions of the lower part of the cliff face.  By moving from 60 meters distance to 10 meters distance and by focusing at the closer distance on the interest points determined at the larger distance, we wished to simulate how a truly autonomous robotic system would approach the cliff face.  Unfortunately, due to a combination of a lack of human foresight in the choice of tripod position and a lack of more advanced software algorithms to mask out the surrounding \& less interesting region (see discussion in Section 4), for one of the two dark spots (Figure 8a), the Cyborg system only found interesting points on the undarkened periphery of the dark \& wet stains. Furthermore, for the other dark spot, the dark spot was spatially complex, being subdivided into several regions, with some green and brown foliage covering part of the mosaic (Figure 8b). Therefore, in both close-up cases the value of the interest mapping is debatable. This interest mapping could be improved in the future, as we discuss in Section 4.2.

The third and final tripod position was at a distance of about 300 meters, near our automobile on the road leading to the cliffs. We were careful to choose a mosaic that properly masked out the grass \& trees as well as the sky. This mosaic ended up being a $1\times4$ mosaic (Figure 6) with a downsampling of 8, which spanned the 10 meter width of the hollow in the cliffs, up to a height of about 30 meters. Again, we were pleased when the system found that the dark \& wet spot near the bottom of the cliff face was the most interesting point. If we had begun the Cyborg Astrobiologist's 2nd mission from this 3rd tripod position, instead of immediately going to the shady tree grove, then
the sum of these results (from the 3rd tripod position to the 1st tripod position; and either discounting or improving the 2nd tripod position) demonstrate that the computer vision algorithms inside the Cyborg Astrobiologist could have served as very reasonable guidance for the geological approach of an autonomous robot towards the cliff face.

 \begin{figure}[h]

 \includegraphics[height=6cm]{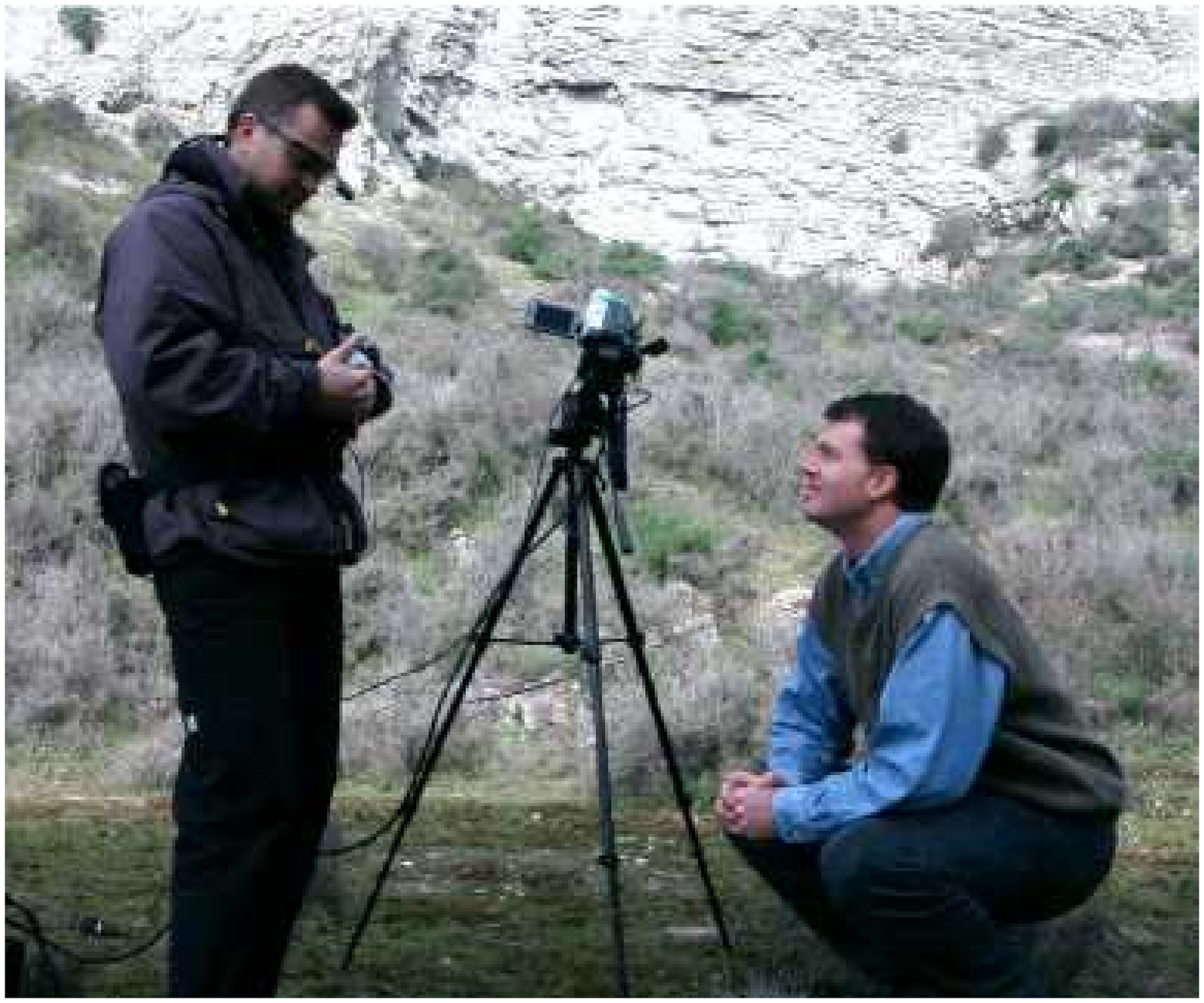}
 \includegraphics[height=6cm]{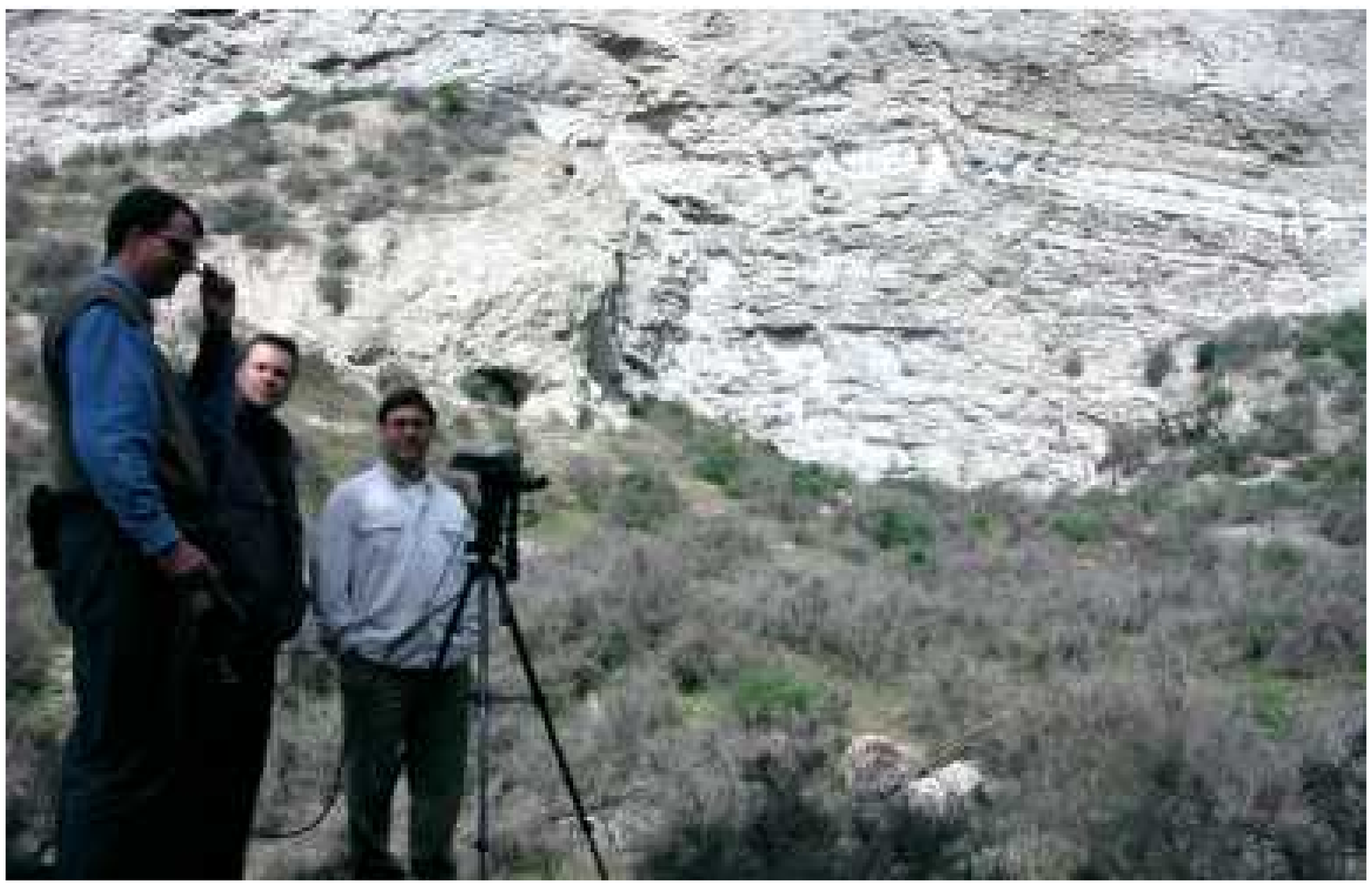}

  \caption{Astrobiologist \& Geologist Orm\"o is wearing the computer part of the Cyborg Astrobiologist system, as Astrobiologist \& Roboticist McGuire looks on. On the right, the team of astrobiologists together for a group pose; during both of these pictures, the robotic part of the Cyborg Astrobiologist was slaving away, acquiring \& processing the images of a $9 \times 4$ vertically oriented mosaic. These pictures were taken during the March 3rd expedition to Rivas Vaciamadrid. Note the absence of the black spots near the bottom of the cliff face; these black spots appeared sometime after this 1st mission and before the 2nd mission. Photos Copyright: D\'iaz Martinez, Orm\"o \& McGuire}

\end{figure}

\section{Results}  
\subsection{Results from the First Geological Field Test}  
\markright{The Cyborg Astrobiologist: Results: First Field Test}

As first observed during the first mission to Rivas on March 3rd, the characteristics of the southward-facing cliffs at Rivas Vaciamadrid consist of mostly tan-colored surfaces, with some white veins or layers, and with significant shadow-causing three-dimensional structure. The computer vision algorithms performed adequately for a first visit to a geological site, but they need to be improved in the future. As decided at the end of the first mission by the mission team, the improvements include:
\begin{spacing}{1}
\begin{trivlist}
\raggedright
\item$\bullet$ Shadow-detection and shadow-interpretation algorithms,
\item$\bullet$ Segmentation of the images based upon microtexture, and
\item$\bullet$ Better methods to control the adaptation of the weighting coefficients determined by the Cortical Interest Map algorithm  (Rae, Fislage \& Ritter 1999), which we used to determine our interest map.
\end{trivlist}
\end{spacing}

 In the last case, we decided that due to the very monochromatic \& slightly-shadowy nature of the imagery, the Cortical Interest Map algorithm non-intuitively decided to concentrate its interest on differences in intensity, and it tended to ignore hue and saturation. 

After the  first geological field test, we spent several months studying the imagery obtained during this mission, and fixing various further problems that were discovered during and after the first mission. Though we had hoped that the first mission to Rivas would have been more like a science mission, in reality it was more of an engineering mission.

\subsection{Results from the Second Geological Field Test}  
\markright{The Cyborg Astrobiologist: Results: Second Field Test}

In Figures 6-8, we show the original mosaics of the intensity sub-images for the three tripod positions from the 2nd mission to Rivas Vaciamadrid.  The computer then found the three most interesting points and repointed the camera at each of these three points, in order to acquire higher-resolution color images. This is an emulation of a mapping \& exploration camera being used first to get the ``lay of the land", after which a higher technology science instrument (i.e.\ a mapping Thermal Emission Spectrometer or M\"ossbauer spectrometer) is trained for more time or with more resources on the most interesting points in the broad landscape. Figures 9-10 show how the computer arrived at its final decision as to where the most interesting points are located, for one of these mosaic images.

In Figure 6, at a distance of 300 meters, the system found the dark wet spot at the bottom of the $4\times1$ mosaic to be the most interesting, the rough texture at the top of the mosaic to be the 2nd most interesting, and a patchy white \& tan colored region in the middle of the mosaic to be the 3rd most interesting. The uncommon map for this mosaic (not shown) shows that the system found the dark \& wet spot to be most interesting largely due the preponderence of pixels of uncommon intensity in this spot. There was significant influence of uncommon saturation pixels for this spot as well.
Furthermore, the uncommonness of the saturation pixels in the dark wet spot (and also the 2nd rough texture interest point) is due to the fact that these saturation pixels are outside the common central peak of the saturation histogram. For our pixel-segmentation algorithm, when the pixels are outside the common central peak, these pixels are often assigned to be in one of the uncommon segments.
Figure 6 might suggest that the algorithm found that the telephone cable (in front of the rough texture at the top of the mosaic) to be the 2nd most interesting point.
However, we find that the rough saturation texture itself is responsible for this region being interesting, and not the telephone cable as one might suspect.


In Figure 7, from the tree grove at a distance of 60 meters, the Cyborg Astrobiologist system found the dark \& wet spot on the right side to be the most interesting, the dark \& wet spot on the left side to be the second most interesting, and the small dark shadow in the upper left hand corner to be the 3rd most interesting. For the first two  interest points (the dark \& wet spots), it is apparent from the uncommon map for intensity pixels in Figure 10 that these points are interesting due to their relatively remarkable intensity values. By inspection of Figure 9, we see that these pixels which reside in the white segment of the intensity segmentation mosaic are unusual because they are a cluster of very dim pixels (relative to the brighter segments, with red, blue and green segment-label colors). Within the dark wet spots,  we observe that these particular points in the white segment of the intensity segmentation in Figure 9 are interesting because they reside in the {\it shadowy} areas of the dark \& wet spots. We interpret the interest in the 3rd interest point to be due to the juxtaposition of the small green plant with the shadowing in this region; the interest in this point is significantly smaller than for the 2 other interest points.

A similar analysis of the segmentation, uncommon and interest maps (not shown here) corresponding to Figures 8a \& 8b, shows, for example:
\begin{spacing}{1}
\begin{trivlist}
\raggedright
\item $\bullet$ The rareness of the two most interesting green- and blue-boxed regions on the periphery of the dark \& wet spot in Figure 8a is due to their being particularly brighter than most of the other pixels.
\item $\bullet$ The peculiarity of the most interesting central white region in Figure 8b is due to its being particularly brighter than most of the other pixels.
\end{trivlist}
\end{spacing}

For both of these cases, the interesting bright regions seem to have a frothy-white crust, more so than the nearby regions which are more tan-colored.

The wearable computer's analysis of the close-up $2\times2$ mosaics of the dark \& wet spots (shown in Figures 8a \& 8b) was somewhat disappointing to us, particularly its analysis of the dark \& wet spot shown in Figure 8a since the mosaic of this dark \& wet spot is less complex than mosaic shown in Figure 8b. In retrospect, given the current advancement of the image-analysis software, we should have taken the mosaic from a distance of 3-5 meters instead of 10 meters, in order to focus {\it only} on the dark \& wet spot, and not upon its periphery. With the current image-analysis software, the computer correctly chose to find the periphery interesting because  the periphery was the smaller region of the image and because the larger and more common dark \& wet spot region was rather homogeneous in its hue \& intensity properties. The dark \& wet spot had particularly diverse properties of saturation; this information could be useful to develop future interest map algorithms for the Cyborg Astrobiologist.

More advanced software could be developed to handle better the close-up real-time interest-map analysis of such imagery as shown in Figures 8a \& 8b. Here are some options to be included in such software development:
\begin{spacing}{1}
\begin{trivlist}
\raggedright
\item $\bullet$ Reintroduce the Cortical Interest Map adaptive weighting algorithm or similar algorithms (i.e., see Chung {\it et al.}\ (2004) for a recent discussion) in order to determine the optimal weighting of the $H$, $S$ \& $I$ uncommon maps for this imagery. This way the system could pay more attention to the Saturation uncommon map if it deems that this map is the salient map.
\item $\bullet$ Enhance the interest map so that it includes maps of edges or other image features in addition to the uncommon maps of hue, saturation \& intensity. This could be useful in searching for interesting boundaries or `contacts' between homogeneous regions, which is a standard technique of field geologists.
\item $\bullet$ Add hardware \& software to the Cyborg Astrobiologist so that it can make intelligent use of its zoom lens. We would need to study \& develop the camera's LANC communication interface, or possibly determine if the zoom lens could be controlled over its Firewire/IEEE1394 communication cable. With accompanying intelligent-zooming software, the system could have corrected the human's mistake in tripod placement and decided to zoom further in, to focus only on the shadowy part of the dark \& wet spot (which was determined to be the most interesting point at a distance of 60 meters), rather than the periphery of the entire dark \& wet spot. 
\item $\bullet$ Enhance the Cyborg Astrobiologist system so that it has a memory of the image segmentations performed at a greater distance or at a lower magnification of the zoom lens. Then, when moving to a closer tripod position or a higher level of zoom-magnification, register the new imagery or the new segmentation maps with the coarser resolution imagery and segmentation maps. Finally, tell the system to mask out or ignore or deemphasize those parts of the higher resolution imagery which were part of the low-interest segments of the coarser, more distant segmentation maps, so that it concentrates on those features that it determined to be interesting at coarse resolution and higher distance. 
\item $\bullet$ Develop true methods of image interpretation, which can define the dark \& wet spot region in Figure 8a as one aspect of the image and the periphery of the dark \& wet spot as another aspect of the image. Then, determine the relationship between these two aspects. Similarly, for Figure 8b, determine, describe and perhaps explain the relationship between the several aspects of this image (dark \& wet spot, frothy-white region to the lower left of the dark \& wet spot; green plant to the left of this frothy white area; dead brown plant to the left of this green plant; and dry, unvegetated areas in the rest of the image).
\end{trivlist}
\end{spacing}

\begin{figure}[h]

\center{\includegraphics[width=4cm]{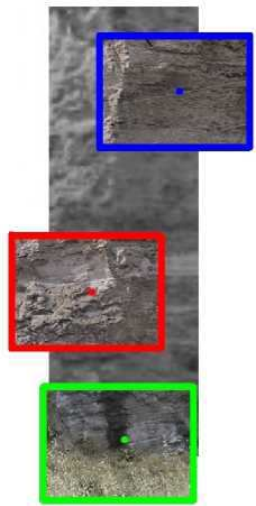}}

\caption{Mosaic image of a four-by-one set of grayscale sub-images acquired by the 
Cyborg Astrobiologist at the end of its second expedition (though we should have acquired it at the beginning of the mission).
The three most interesting points were subsequently revisited by the
camera in order to acquire full-color higher-resolution images of
these points-of-interest. 
 The colored points and rectangles represent the points that the Cyborg
 Astrobiologist determined (on location) to be most interesting;
{\it green} is most interesting, {\it blue} is second most interesting, and {\it red} is third most interesting.
    The images were taken and processed in real-time between 4:59PM and 5:02PM local time
on 11 June 2004 about 300 meters from some gypsum-bearing southward-facing
cliffs near the ``El Campillo" lake of the Madrid southeast regional park
outside of Rivas Vaciamadrid. The most interesting point ({\it green} box)
corresponds to the wet \& darkened area near the bottom of the cliffs; this interest point could have been used to guide the geological approach to 30 meters and then to 10 meters  to the interesting dark \& wet spot by the Cyborg Astrobiologist earlier in the day, as represented in the next two figures. 
 }
 \end{figure}

 \begin{figure}[h]

\center{\includegraphics[width=13.5cm]{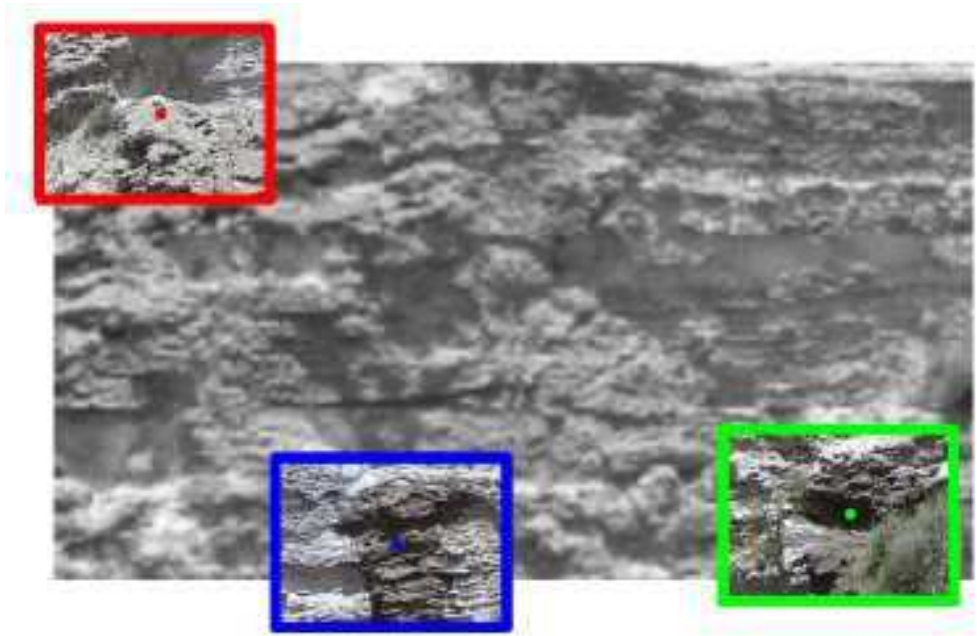}}

\caption{Mosaic image of a three-by-four set of grayscale sub-images acquired by the 
Cyborg Astrobiologist at the beginning of its second expedition.
The three most interesting points were subsequently revisited by the
camera in order to acquire full-color higher-resolution images of
these points-of-interest.  
 The colored points and rectangles represent the points that the Cyborg
 Astrobiologist determined (on location) to be most interesting;
{\it green} is most interesting, {\it blue} is second most interesting, and {\it red} is third most interesting.
    The images were taken and processed in real-time between 1:25PM and 1:35PM local time
on 11 June 2004 about 60 meters from some gypsum-bearing southward-facing
cliffs near the ``El Campillo" lake of the Madrid southeast regional park
outside of Rivas Vaciamadrid. See Figures 9 and 10 for some details about the real-time image processing that was done in order to determine the location of the interest points in this figure.
 }

 \end{figure}

 \begin{figure}[t]

\center{ \includegraphics[width=11.5cm]{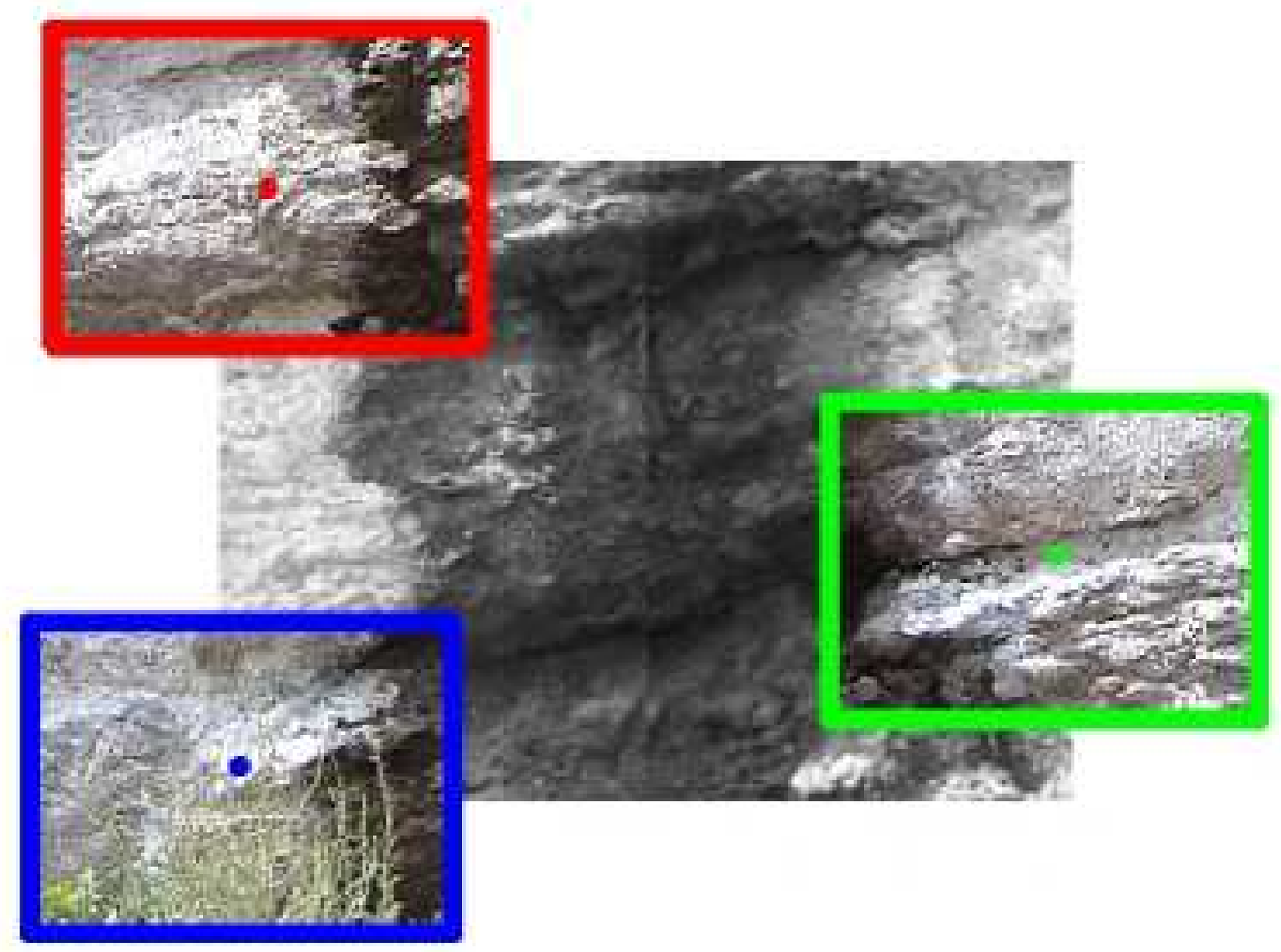}}
\center{a)}
 \center{\includegraphics[width=11.5cm]{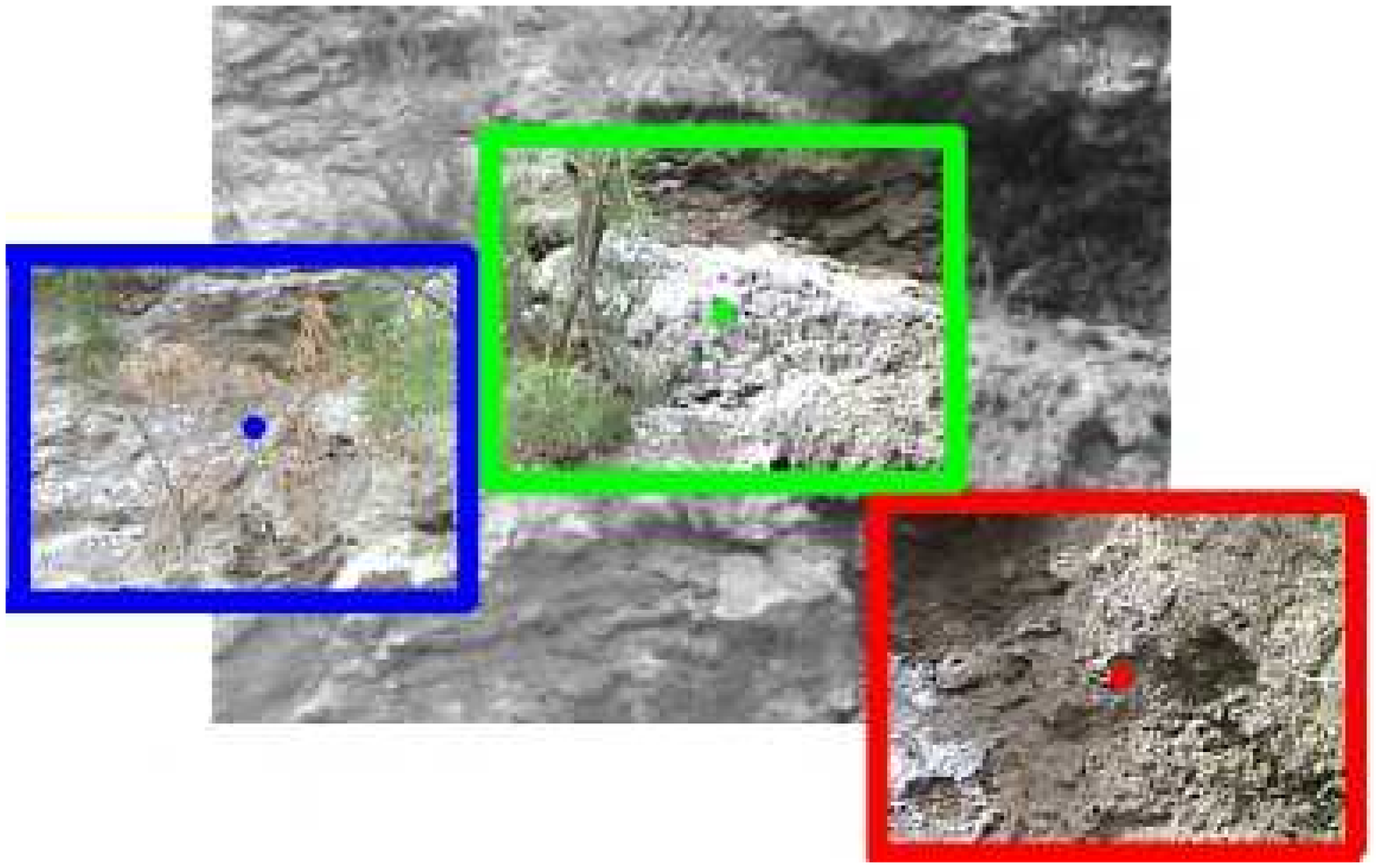}}
\center{b)}

\caption{Two different mosaic images of a two-by-two set of grayscale sub-images acquired by the 
Cyborg Astrobiologist in the middle of its second expedition.
The three most interesting points were subsequently revisited by the
camera in order to acquire full-color higher-resolution images of
these points-of-interest.  
 The colored points and rectangles represent the points that the Cyborg
 Astrobiologist determined (on location) to be most interesting;
{\it green} is most interesting, {\it blue} is second most interesting, and {\it red} is third most interesting.
    The images were taken and processed in real-time between 3:27PM and 
at 4:05PM local time on 11 June 2004 about 10 meters from some gypsum-bearing southward-facing
cliffs near the ``El Campillo" lake of the Madrid southeast regional park
outside of Rivas Vaciamadrid. The upper mosaic (a) represents the Cyborg's
study of the left-most dark \& wet spot of Figure 7, and the lower mosaic (b) represents
the Cyborg's study of the right-most dark \& wet spot of Figure 7. 
 }

 \end{figure}

\section{Discussion \& Conclusions}

   Both the geologists on our team concur with the judgement of the Cyborg Astrobiologist software system, that the two dark \& wet spots on the cliff wall were the most interesting spots during the second mission. However, the two geologists also state that this largely depends on the aims of study for the geological field trip; if the aim of the study is to search for hydrological features, then these two dark \& wet spots are certainly interesting; however if faulting between layers is of interest, then the dark \& wet spots would not retain as much interest. One question which we have thus far left unstudied is ``What would the Cyborg Astrobiologist system have found interesting during the second mission if the two dark \& wet spots had not been present during the second mission?'' It is possible that it would again have found some dark shadow particularly interesting, but with the improvements made to the system between the first and second mission, it is also possible that it could have found a different feature of the cliff wall more interesting.

\subsection{Thought Experiment: Blueberry Fields in the Meridiani Planum}
\markright{The Cyborg Astrobiologist: Discussion: Blueberry Fields at Meridiani}
  One of the remarkable discoveries made by the MER team this year is the existence of small, dark \& round pebbles (a.k.a. `blueberries') in many places near the landing site of the Opportunity rover Meridiani Planum. The Opportunity rover was sent to Meridiani Planum because this was the main site on Mars that had been observed by the Thermal Emission Spectrometer on the Mars Global Surveyor orbiter to have a large quantity of gray hematite (Christensen {\it et al.}\ 2000). Gray hematite is an iron-bearing mineral which often forms in the presence of water, particularly as a precipitate in cold standing water (Glotch {\it et al.}\ 2004). The so-called blueberries that Opportunity discovered at Meridiani Planum were also observed to have significant hematite in them, based upon studies from both of the mini-TES and M\"ossbauer spectrometer instruments onboard the Opportunity rover (Squyres, unpublished\footnote{lecture notes from the course on the MER rovers from Astrobiology Summer School {\it ``Planet Mars''}, Universidad Internacional Menendez Pelayo Santander, Spain}, 2004). There is significant excitement in the planetary geology community about this discovery, partly due to the existence of a place on Earth (Utah) where similar blueberries have been formed (Chan {\it et al.}\ 2004; Squyres \& Athena Science Team, 2004; Moore 2004; Catling 2004).

  Such a geological phenomenon suggests a thought experiment that could be applied to our Cyborg Astrobiologist system as it currently exists: ``If the Cyborg Astrobiologist system was taken to a field covered with Meridiani/Utah blueberries, would it find the blueberries to be interesting?'' The brief, unqualified answer to this thought experiment is: ``No''. If the blueberries covered the landing site, so that their coverage exceeded some percentage (assume for argument's sake that the areal coverage of blueberries at the location of the Cyborg Astrobiologist is 80\%), then an image mosaic taken by the Cyborg Astrobiologist would be dominated by dark blueberries. Therefore, the Cyborg Astrobiologist's uncommon map algorithm, as it now exists, would likely search the remaining 20\% of the area for the smallest color class of pixels, and it would ignore the blueberries.

   However, if the Cyborg Astrobiologist was allowed to approach the field of blueberries from a distance (i.e., from orbit or from a neighboring area without blueberries), then it is probable that the Cyborg Astrobiologist would at some point find the blueberries to be new and interesting. This is because blueberries don't cover all of Mars, and so at some spatial scale, they only have a small areal coverage. So the qualified answer to the thought experiment is: ``Yes''.

   Furthermore, we could program more advanced renditions of the software inside of the Cyborg Astrobiologist so that it could find the blueberries interesting even if the Cyborg Astrobiologist was not allowed to approach the blueberry field from a distance. Such algorithm enhancements could include the texture \& edge information suggested above, as well as a possible algorithm to search for uncommon shapes.
With spectrometer information (i.e., visible/near-infrared, mid-infrared, Raman, and/or M\"ossbauer) and with biased searching for
particularly-interesting spectra, textures, rock shapes and/or rock juxtapositions, the odds of success could also be maximized, either in hunting for blueberries or in serendipitously finding such blueberries.

These advances in our system could take some years to implement, but we will surely implement some of them in the near-term. In the meantime, as discussed above, the valued concept of a geological approach will remain our solution to the `blueberry-field' thought experiment. Nevertheless, such a thought experiment does point out a significant inadequacy of our simple computer vision algorithm which searches for small regions of the color imagery, as discussed in the preceding three paragraphs. Such simple algorithms need to be enhanced and brought into relation with other algorithms, and perhaps with a visual memory of past imagery (i.e., the VAMPIRE project (Wrede {\it et al.}\ 2004)).

A visual memory would also be useful to record and track changes in the areal coverage of components in the scenery, as mosaics or images are acquired from different camera positions or for different outcrops. In a different outcrop, the dark wet spots might dominate the whole outcrop, or the whitish color from the gypsum might not dominate the outcrop. The visual memory could keep track of outcrops studied in the recent past, and inform the user when an uncommon area of a particular image or mosaic is truly novel in areal coverage, when compared with an area of similar color or texture characteristics in prior imagery.

\subsection{Outlook}
\markright{The Cyborg Astrobiologist: Outlook}

   The NEO programming for this Cyborg Astrobiologist/Geologist project was initiated with the SONY Handycam in April 2002. The wearable computer arrived in June 2003, and the head mount\-ed display arrived in November 2003. We now have a reliably functioning human and hardware and software Cyborg Geologist system, which is partly robotic with its Pan Tilt camera mount. This robotic extension allows the camera to be pointed repeatedly, precisely \& automatically in different directions. 

Prior to each of the two field-trips to Rivas Vaciamadrid, the Cyborg Geologist system was tested on the grounds of our Spanish research institute (CAB). These tests were mostly on a  wide prairie field: by winter, this prairie field was plowed and full of 3-10 cm diameter stones of different, dull colors; and by summer, this prairie field was full of prairie grasses and differently-colored wildflowers. In these nearby field-trips, the Cyborg Geologist has been providing us with constructive criticism of the basic capabilities of our image-segment\-ation and interest-map algorithms.

 As of our initial field-trips to Rivas, the Cyborg Geologist system has accompanied two Human field geologists to a site of geological interest (i.e. layered rock outcrops) for more advanced testing of these computer-vision algorithms.  We plan several more such trips in the coming months.

      Based upon the performance of the Cyborg Astrobiologist system during the 1st mission to Rivas in March 2004 on the outcropping cliffs near Rivas Vaciamadrid, we have decided that the system was paying too much attention to the shadows made by the 3D structure of the cliffs. We hope to improve the Cyborg Astrobiologist system in the next months in order to detect and to pay less attention to shadows. We also hope to upgrade our system to include: image-segmentation based upon micro-texture; and adaptive methods for summing the uncommon maps in order to compute the interest map.

      Based upon the significantly-improved performance of the Cyborg Astrobiologist system during the 2nd mission to Rivas in June 2004, we conclude that the system now is debugged sufficiently so as to be able to produce studies of the utility of particular computer vision algorithms for geological deployment in the field. We have outlined some possibilities for improvement of the system based upon the second field trip, particularly in the improvement in the systems-level algorithms needed in order to more intelligently drive the approach of the Cyborg or robotic system towards a complex geological outcrop. These possible systems-level improvements include: a better interest-map algorithm, with adaptation and more layers; hardware \& software for intelligent use of the camera's zoom lens; a memory of the image segmentation performed at greater distance or lower magnification of the zoom lens; and high-level image-interpretation capabilities.

Now that we have demonstrated that this software and hardware in the Cyborg Astrobiologist system can function for developing and testing computer-vision algorithms for robotic exploration of a geological site, we have some decisions to make as to future directions for this project, options for these future directions include:
\begin{spacing}{1}
\begin{trivlist}
\raggedright
\item $\bullet$ Performing further offline analysis and algorithm-development for the imagery obtained at Rivas Vaciamadrid: several of the parameters of the algorithms need testing for their optimality, and further enhancements of the algorithms could be made.
\item $\bullet$ Optimizing the image-processing \& robotic-control code for the current Cyborg Astrobiologist system for speed and memory utilization. 
\item $\bullet$ Further testing of the existing Cyborg geological exploration system at other geological sites with different types of imagery.
\item $\bullet$ Speeding up the algorithm development by changing the project from being partly a hardware project with cameras and pan-tilt units and fieldwork to being entirely a software project without robotically-obtained image mosaics and without robotic interest-map pointing;
with such a change in focus, our algorithms could be significantly enhanced by studying many more types of imagery: for example, from human geologist field studies on the Earth, from robotic geologist field studies on Mars, and from orbiter or flyby studies of our solar system's moons.
\end{trivlist}
\end{spacing}

\section{Acknowledgements}

P. McGuire, J. Orm\"o and E. D\'iaz Mart\'inez would all like to thank the Ramon y Cajal Fellowship program in Spain, as well as certain individuals for assistance or conversations: 
Virginia Souza-Egipsy, Eduardo Sebasti\'an Mart\'inez, Mar\'ia Paz Zorzano Mier, Carmen C\'ordoba Jabonero, Kai Neuffer, Antonino Giaquinta, Fernando Camps Mart\'inez, Alain Lepinette Malvitte, Josefina Torres Redondo, V\'ictor R. Ruiz, Julio Jos\'e Romeral Planell\'o, Gemma Delicado, Jes\'us Mart\'inez Fr\'ias, Irene Schneider, Gloria Gallego, Carmen Gonz\'alez, Ramon Fern\'andez, Coronel Angel Santamaria, Carol Stoker, Paula Grunthaner, Maxwell D. Walter, Fernando Ayll\'on Quevedo, Javier Mart\'in Soler, Juan P\'erez Mercader, J\"org Walter, Claudia Noelker, Gunther Heidemann, Robert Haschke, Robert Rae, Jonathan Lunine, and two anonymous referees. The work by J. Orm\"o was partially supported by a grant from the Spanish Ministry for Science and Technology (AYA2003-01203). The equipment used in this work was purchased by grants to our Center for Astrobiology from its sponsoring research organizations, CSIC and INTA.


\appendix
\section{Appendix: Software Implementation Details}

Hue, Saturation and Intensity are computed in NEO by the standard {\it HSI} subroutine, which uses the HSI model described by Foley {\it et al.}\ (1990). Hue ($H$) corresponds roughly to the color of a pixel; saturation ($S$) corresponds roughly to the purity of the hue; and intensity corresponds to the total brightness of the pixel.  Then each of these 3 $H,S,I$ image slices is downsampled by a given factor $D$ in each direction ($D$ is typically 2 or 4 or 8) in order to conserve memory.

An important but very technical detail in our implementation is that due to the common juxtaposition of pixels of widely different Hue ($H$) in the Hue slice (especially for image regions of low Saturation), we needed to perform a type of median filter on the Hue slice in non-overlapping $D \times D$ neighborhoods prior to the downsampling. Such median filtering prior to downsampling was not as important for $S$ and $I$ as it was for $H$.

The median-filtered and downsampled images are finally combined together into a quasi-mosaic of dimensions $M \times N$ subimages by simply butting the subimages together. We do not perform any mathematical computations to ensure that the edges of the mosaic subimages match well with the edges of the neighboring subimages.

Following Haralick, Shanmugan \& Dinstein (1973), we developed custom 2D `co-occurrence' histogramming C code in NST for pairs of pixels of a given separation and orientation, using min/max stretching for preprocessing prior to creating the histograms. For these studies, we chose to histogram pixels separated by 1 pixel width and which are horizontal neighbors, but these 2D histograms are separately made (Figure 9) for each of the Hue, Saturation and Intensity image slices. For other studies, we have chosen, for example, to make eight simultaneous and separate 2D histograms for pixels separated by 5 pixels and by 1 pixel, and with pixel orientations of 0, 45, 90, and 135 degrees. Perhaps in the future, we can implement and test something akin to a wavelet analyzer of texture.

With the three 2D histograms (for $H$, $S$ \& $I$), we use the standard {\it locate\_peaks} peak finder in NEO for searching for the eight largest peaks in the 2D histogram. After {\it locate\_peaks} finds a peak, it masks out histogram bins in a disk of a given radius $w$ around that peak, and then it searches the remaining histogram bins for the subsequent peaks of smaller amplitude. The values of $w$ were chosen by trial and error based upon previous field work, with the values of $w$ for $H$ \& $S$ being 15 stretched units, and the value of $w$ for $I$ being 20 stretched units. This standard {\it locate\_peaks} peak finder may not be very advanced, but it is rather fast and somewhat robust. This robustness is somewhat crucial, since the structure of the 2D Haralick co-occurrence histograms often have structures that might confound a less robust but more advanced peak finder. These 2D histogram structures in the Haralick co-occurrence method, which might confound a more advanced peak finder, include: significant variation in the choppiness (or lack of smooth variation) of the co-occurrence histogram as well as significant variation in the width of the true peaks in the co-occurrence histogram. Furthermore, the co-occurrence histograms often have only one true peak, which is somewhat larger in width than the chosen value of $w$. In this case of only a single true peak, the system will find the first true peak, and it will then find the remaining 7 peaks to be the highest points in the histogram that are more distant than $w$ histogram bins from the true peak. This may seem somewhat artificial, but in practice, it is a decent method to search for those pixels in the image which are truly unusual and interesting, which is our main task in this project. Nonetheless, this peak finding algorithm does merit significant study and possible revision in future work.

After finding the 8 peaks in a 2D histogram of image pixel-pairs, we reordered the peaks, ranking the histogram peaks in the order of the number of original-image pixel-pairs within their $w$ neighborhoods (each bin in the 2D histogram counts the number of original-image pixel-pairs that have the corresponding values of the ordinate \& abscissa). This reordering may seem artificial, but it was necessary given the nature of the {\it locate\_peaks} peak finding algorithm that we used. To make the possible artificiality clear: the peaks were found based upon the largest remaining single pixel values in the 2D co-occurrence histograms, but the peaks were ranked based upon the numbers of pixels in their $w$ neighborhoods. An alternative is to use a more coherent scheme which finds the 8 pre-ranked locations of the $w$ neighborhoods with largest numbers of pixels. However, we defer this $w$ neighborhood peak-finding algorithm development for the future since the present system seems to work tolerably well and quickly, and since the artifacts we have observed from using the current scheme of peak finding and peak ordering have not yet suggested the urgency of such an upgraded algorithm.

After the 2D histogramming and peak finding, the images are segmented by our NEO $prog\_unit$ implementation (custom {\it C} code) of the Haddon \& Boyce (1990) method. We throw out pixels that have possible conflicts, unlike Haddon \& Boyce's more careful work.  We handle with some care the pixels on the edges of the subimages in the quasi-mosaic. However, in practice, these edge pixels cause the most problems with the image segmentation when the motor for the Pan-Tilt unit is taking mosaicking steps that do not match well the optical zoom factor of the camera, so that the edges of the subimages are easily visible in the quasi-mosaic. A secondary source of subimage edge problems for this image segmentation occurs when there are cumulus clouds in the sky, occasionally occulting the Sun, which causes some of the subimages to have different lighting properties than their neighbors.  If a pixel and its horizontal neighbor have values that sets it in one peak's neighborhood (i.e., peak number $s$ of the 8 ranked peaks of that 2D histogram), then the pixel is said to be in segment $s$ of that image. The segmentation maps for that image are therefore colored by 8 different colors, with red for the `most common' segment $s=1$ (which has the most pixels in it), blue for 2nd `most common' segment $s=2$, etc. See Figure 9 for an example. 

\makeatletter	
\renewcommand{\@biblabel}[1]{}
\makeatother

\end{spacing}

\markright{The Cyborg Astrobiologist: First Field Experience}

 \begin{figure}[t]

\center{\includegraphics[width=15.0cm]{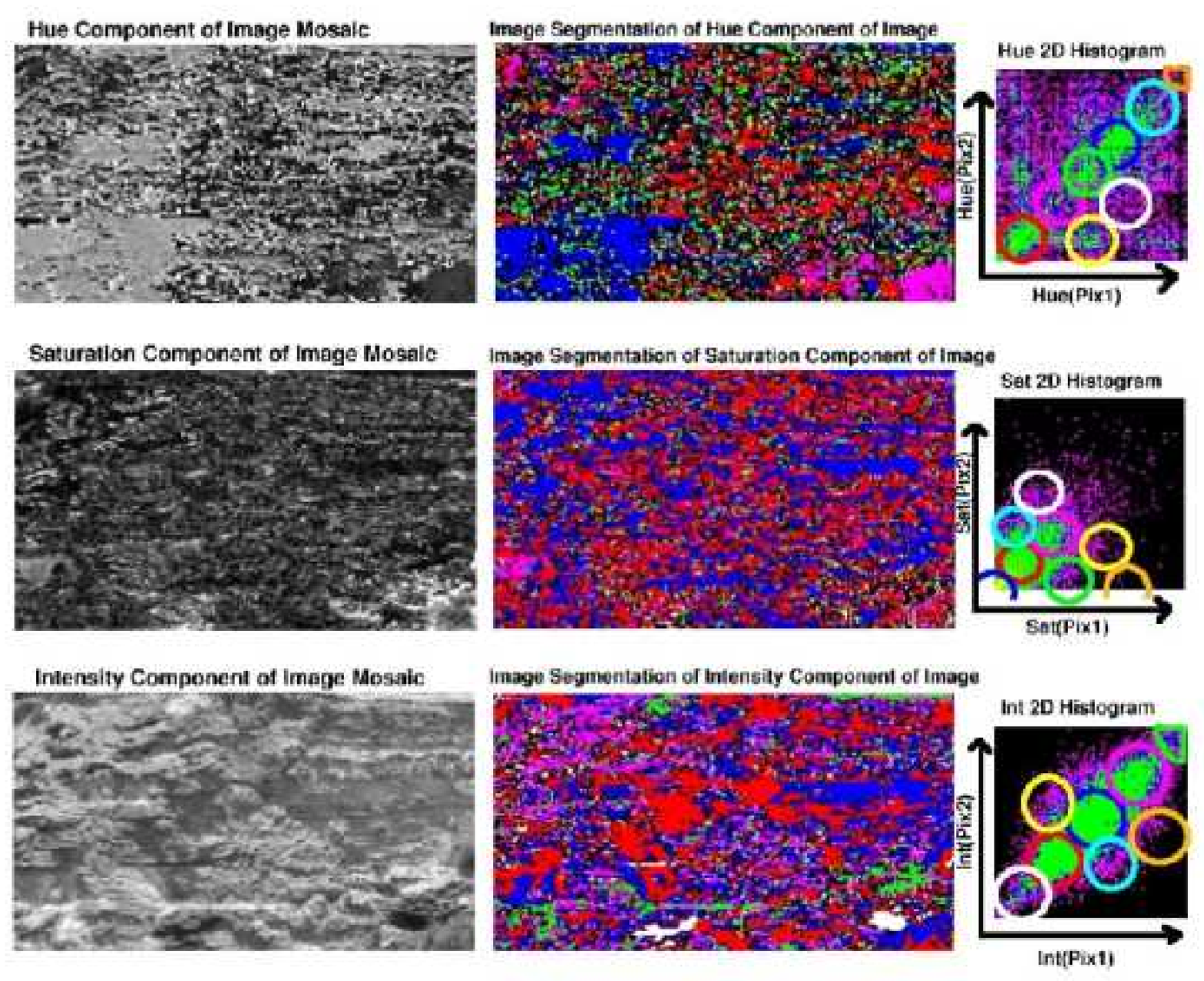}}

\caption{
   In the middle column, we show the three image-segmentation maps computed in real-time by the Cyborg Astrobiologist system, based upon the original Hue, Saturation \& Intensity ($H$,$S$ \& $I$) mosaics in the left column and the derived 2D co-occurrence histograms shown in the right column. The wearable computer made this and all other computations for the original $3\times4$ mosaic ($108\times192$ pixels, shown in Figure 7) in about 2 minutes after the initial acquisition of the mosaic sub-images was completed.
   The colored regions in each of the three image-segmentation maps
correspond to pixels \& their neighbors in that map that have similar
statistical properties in their two-point correlation values, as shown by the circles of corresponding colors in the 2D histograms in the column on the right. 
         The RED-colored regions in the segmentation maps correspond to the mono-statistical
              regions with the largest area in this mosaic image;
              the RED regions are the least "uncommon" pixels in the
              mosaic.
         The BLUE-colored regions correspond to the mono-statistical
              regions with the 2nd largest area in this mosaic image;
              the BLUE regions are the 2nd least "uncommon" pixels in the
              mosaic.
         And similarily for the PURPLE, GREEN, CYAN, YELLOW, WHITE, and ORANGE.
         The pixels in the BLACK regions have failed to be segmented by the segmentation algorithm.
}

 \end{figure}

 \begin{figure}[t]

\center{\includegraphics[height=20.0cm]{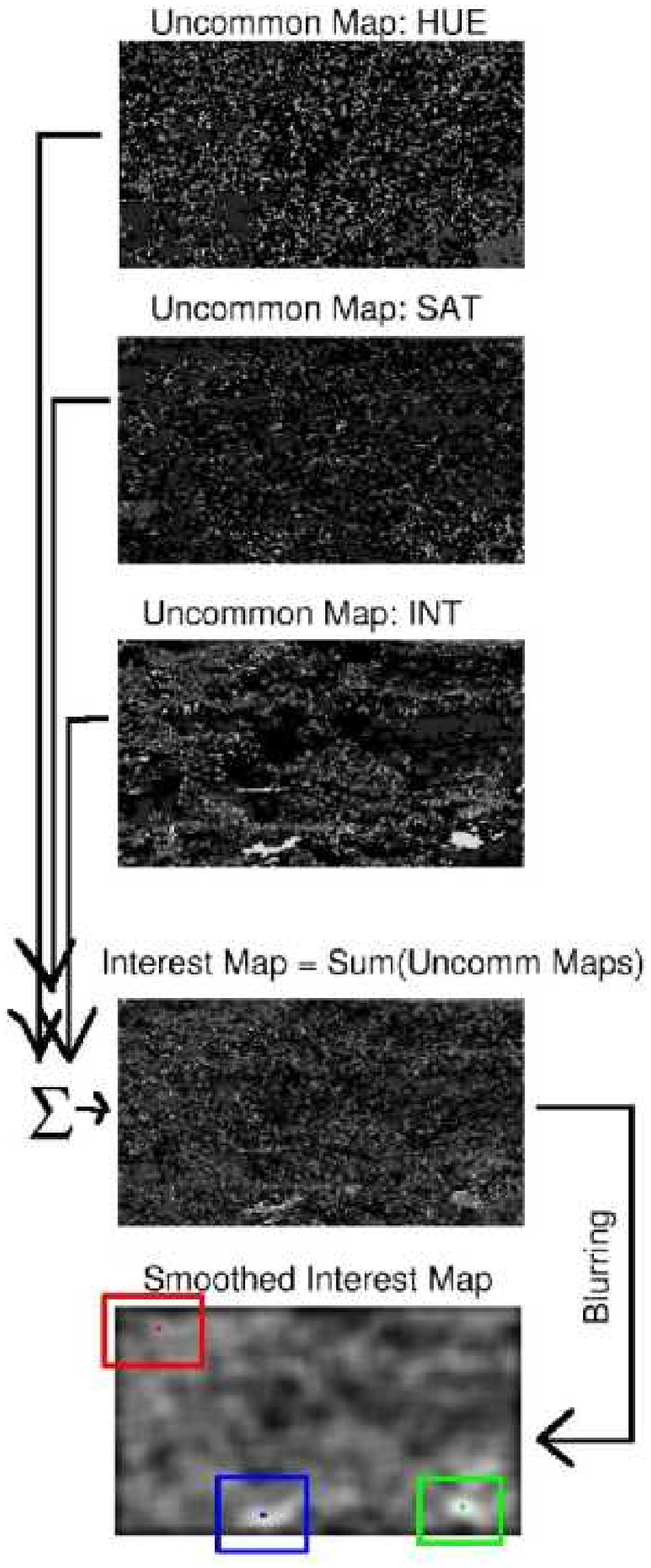}}

\caption{
  These are the uncommon maps for the mosaic shown in Figure 7,  based on the region sizes determined by the image-segmentation algorithm shown in  Figure 9.
  Also shown is the interest map, i.e., the unweighted sum of the three uncommon maps. We blur the original interest map before
determining the ``most interesting" points. These ``most interesting" points
are then sent to the camera's Pan/Tilt motor in order to acquire and
save-to-disk 3 higher-resolution RGB color images of the small areas in
the image around the interest points (see Figure 7).
   Green is the most interesting point. Blue is 2nd most interesting.
And Red is 3rd most interesting.
}

 \end{figure}


\begin{thebibliography}{}
\markright{The Cyborg Astrobiologist: References}


\bibitem{}
Apostolopoulos, D., Wagner, M.D., Shamah, B., Pedersen, L.,  Shillcutt, K. \& Whittaker, W.L. (2000). 
Technology and Field Demonstration of Robotic Search for Antarctic Meteorites.
{\it International Journal of Robotics Research} {\bf 19}(11), pp. 1015-1032. 


\bibitem{}
Arvidson, R.~E., Anderson, R.~C., Haldemann, A.~F.~C., Landis, 
G.~A., Li, R., Lindemann, R.~A., Matijevic, J.~R., Morris, R.~V., Richter, 
L., Squyres, S.~W., Sullivan, R.~J. \& Snider, N.~O.\ (2003). Physical 
properties and localization investigations associated with the 2003 Mars 
Exploration rovers.  {\it Journal of Geophysical Research (Planets)}  {\bf 108}, 11-1. 

	\bibitem{} 
Arvidson, R.~E., Anderson, R.~C., Bartlett, P., Bell, J.~F., Blaney, D., 
Christensen, P.~R., Chu, P., Crumpler, L., Davis, K., Ehlmann, B.~L., 
Fergason, R., Golombek, M.~P., Gorevan, S., Grant, J.~A., Greeley, R., 
Guinness, E.~A., Haldemann, A.~F.~C., Herkenhoff, K., Johnson, J., Landis, 
G., Li, R., Lindemann, R., McSween, H., Ming, D.~W., Myrick, T., Richter, 
L., Seelos, F.~P., Squyres, S.~W., Sullivan, R.~J., Wang, A. \& Wilson, 
J.\ (2004). Localization and Physical Properties Experiments Conducted by 
Spirit at Gusev Crater.\  {\it Science} {\bf 305}, 821-824. 



\bibitem{}
Batavia, P.H. \&  Singh, S. (2001). Obstacle detection using adaptive color segmentation and color stereo homography.
{\it Proceedings of the IEEE Conference on Robotics and Automation}, Seoul, Korea, pp. 705-710.

\bibitem{}
Batlle, J., Casals, A., Freixenet, J. \& Mart\'i, J. (2000). A review for recognizing natural objects in colour images of outdoor scenes. {\it Image and Vision Computing} {\bf 18}, pp. 515-30.

\bibitem{} 
Bell, J.~F., Squyres, S.~W., Herkenhoff, K.~E., Maki, J.~N., Arneson, 
H.~M., Brown, D., Collins, S.~A., Dingizian, A., Elliot, S.~T., Hagerott, 
E.~C., Hayes, A.~G., Johnson, M.~J., Johnson, J.~R., Joseph, J., Kinch, K., 
Lemmon, M.~T., Morris, R.~V., Scherr, L., Schwochert, M., Shepard, M.~K., 
Smith, G.~H., Sohl-Dickstein, J.~N., Sullivan, R.~J., Sullivan, W.~T. \&
Wadsworth, M.\ (2003). Mars Exploration Rover Athena Panoramic Camera 
(Pancam) investigation.  {\it Journal of Geophysical Research (Planets)}  
{\bf 108}, 4-1. 

\bibitem{} 
Bell, J.~F., Squyres, S.~W., Arvidson, R.~E., Arneson, H.~M., Bass, D., 
Blaney, D., Cabrol, N., Calvin, W., Farmer, J., Farrand, W.~H., Goetz, W., 
Golombek, M., Grant, J.~A., Greeley, R., Guinness, E., Hayes, A.~G., 
Hubbard, M.~Y.~H., Herkenhoff, K.~E., Johnson, M.~J., Johnson, J.~R., 
Joseph, J., Kinch, K.~M., Lemmon, M.~T., Li, R., Madsen, M.~B., Maki, 
J.~N., Malin, M., McCartney, E., McLennan, S., McSween, H.~Y., Ming, D.~W., 
Moersch, J.~E., Morris, R.~V., Noe Dobrea, E.~Z., Parker, T.~J., Proton, 
J., Rice, J.~W., Seelos, F., Soderblom, J., Soderblom, L.~A., 
Sohl-Dickstein, J.~N., Sullivan, R.~J., Wolff, M.~J. \& Wang, A.\ (2004). 
Pancam Multispectral Imaging Results from the Spirit Rover at Gusev 
Crater.\  {\it Science} {\bf 305}, 800-807. 

\bibitem{} 
Cabrol, N.~A., Glenister, B., Chong, G., Herrera, C., Jensen, A., Pereira, 
M., Stoker, C.~R., Grin, E.~A., Landheim, R., Thomas, G., Golden, J., 
Saville, K., Ludvigson, G. \& Witzke, B.\ (2001). Nomad Rover Field 
Experiment, Atacama Desert, Chile 2. Identification of paleolife evidence 
using a robotic vehicle: Lessons and recommendations for a Mars sample 
return mission.\  {\it Journal of Geophysical Research} {\bf 106} E4, pp. 7807-7811. 

\bibitem{}
Calvo, J.P.,  Alonso, A.M. \& Garcia del Cura, A.M. (1989). Models of
Miocene marginal lacustrine sedimentation in response to varied
depositional regimes and source areas in the Madrid Basin (central
Spain). {\it Paleogeogr. Paleoclim, Paleoecol.} {\bf 70}, 199-214.




\bibitem{}
Catling, D.C. (2004). Planetary Science: On Earth, as it is on Mars? {\it Nature} {\bf 429}, 707-708.

\bibitem{}
Chan, M.A., Beitler, B., Parry, W.T., Orm\"o, J. \& Komatsu, G. (2004). 
A possible terrestrial analogue for haematite concretions on Mars.
{\it Nature} {\bf 429}, 731-734. 

\bibitem{}
Cheeseman, P., Kelly, J., Self, M., Stutz, J., Taylor, W., \& Freeman, D. (1988). AutoClass: A Bayesian Classification System.  {\it Proceedings of the Fifth International Conference on Machine Learning}, Ann Arbor, MI. June 12-14 1988. Morgan Kaufmann Publishers, San Francisco, pp. 54-64.

\bibitem{}
Cheeseman, P. \& Stutz, J. (1996). Bayesian Classification (AutoClass): Theory and Results, {\it Advances in Knowledge Discovery and Data Mining}, Usama M. Fayyad, Gregory Piatetsky-Shapiro, Padhraic Smyth, \& Ramasamy Uthurusamy, Eds. AAAI Press/MIT Press.

\bibitem{} 
Christensen, P.~R., Bandfield, J.~L., Clark, R.~N., Edgett, K.~S., 
Hamilton, V.~E., Hoefen, T., Kieffer, H.~H., Kuzmin, R.~O., Lane, M.~D., 
Malin, M.~C., Morris, R.~V., Pearl, J.~C., Pearson, R., Roush, T.~L., Ruff, 
S.~W. \& Smith, M.~D.  (2000). Detection of crystalline hematite mineralization on Mars by the Thermal Emission Spectrometer: Evidence for near-surface water. {\it Journal of Geophysical Research} {\bf 105}, 9623-9642. 

\bibitem{} 
Christensen, P.~R., Ruff, S.~W., Fergason, R.~L., Knudson, A.~T., Anwar, 
S., Arvidson, R.~E., Bandfield, J.~L., Blaney, D.~L., Budney, C., Calvin, 
W.~M., Glotch, T.~D., Golombek, M.~P., Gorelick, N., Graff, T.~G., 
Hamilton, V.~E., Hayes, A., Johnson, J.~R., McSween, H.~Y., Mehall, G.~L., 
Mehall, L.~K., Moersch, J.~E., Morris, R.~V., Rogers, A.~D., Smith, M.~D., 
Squyres, S.~W., Wolff, M.~J. \& Wyatt, M.~B.\ (2004). Initial Results from 
the Mini-TES Experiment in Gusev Crater from the Spirit Rover.\  {\it 
Science} {\bf 305}, 837-842. 


\bibitem{}
Chung, A.J., Deligianni, F., Hu, X.P. \& Yang, G.Z. (2004).  Visual feature extraction via eye tracking for saliency driven 2D/3D registration. {\it Conference on Eye Tracking Research and Applications (ETRA04)}, San Antonio, Texas, USA, pp. 49-54.


\bibitem{}
Corsetti, F.A. \& Storrie-Lombardi, M.C. (2003). Lossless Compression of Stromatolite Images: a Biogenicity Index? {\it Astrobiology}, {\bf 3}(4), pp. 649-655.






\bibitem{} 
Crisp, J.~A., Adler, M., Matijevic, J.~R., Squyres, S.~W., Arvidson, R.~E.
\& Kass, D.~M.\ (2003). Mars Exploration Rover mission.\  {\it Journal of 
Geophysical Research (Planets)} {\bf 108}, 2-1. 

\bibitem{} 
Crumpler, L., Cabrol, N., Des Marais, D., Farmer, J., Golombek, M., Grant, 
J., Greeley, R., Grotzinger, J., Haskin, L., Arvidson, R., Squyres, S., 
Learner, Z., Li, R., Madsen, M.~B., Malin, M., Payne, M., Parker, T., 
Seelos, F., Sims, M., de Souza, P., Wang, A., Weitz, C. \& The Athena 
Science Team (2004). MER Field Geologic Traverse in Gusev Crater, Mars: 
Initial Results From the Perspective of Spirit.\  {\it Lunar and Planetary 
Institute Conference Abstracts} {\bf 35}, 2183. 

\bibitem{}
Foley, J.,  van Dam, A., Fiener, S. \& Hughes, J. (1990). {\it Computer Graphics: Principles and Practice} (2nd Edition) Addison-Wesley. 

\bibitem{}
F{\"o}rstner, W. (1986). A feature based algorithm for image matching. {\it International Archives of Photogrammetry and Remote Sensing}, {\bf 26}(3), pp. 150-166.

\bibitem{}
F{\"o}rstner, W. \& G{\"u}lch, E. (1987). A Fast Operator for Detection and Precise Location of Distinct Points, Corners and Centres of Circular Features. {\it Proc. Intercommission Conference on Fast Processing of Photogrammetric Data}, Interlaken, Switzerland. pp. 281-305.


\bibitem{}
Freixenet, J., Mu\~noz, X., Mart\'i, J. \& Llad\'o, X. (2004). Color Texture Segmentation by Region-Boundary Cooperation. {\it Computer Vision -- ECCV 2004, Eighth European Conference on Computer Vision, Proceedings, Part II, Lecture Notes in Computer Science} Springer, Prague, Czech Republic, Eds. Tom\'as Pajdla, Jir\'i Matas, {\bf 3022}, 250-261. 
Also available in the {\it CVonline} archive:\\ http://homepages.inf.ed.ac.uk/rbf/CVonline/LOCAL\_COPIES/FREIXENET1/eccv04.html

\bibitem{} 
Glotch, T.~D., Christensen, P.~R., Wyatt, M.~B., Bandfield, J.~L., Graff, 
T.~G., Rogers, D., Ruff, S.~W., Hayes, A.~G., Morris, R.~V., Farrand, W., 
Calvin, W., Moersch, J.~E., Ghosh, A., Johnson, J.~R., Fallacaro, A., 
Blaney, D., Squyres, S.~W., Bell, J.~F., Klingelh{\" o}fer, G., Souza, P. 
\& The Athena Science Team (2004). Hematite at Meridiani Planum: Detailed 
Spectroscopic Observations and Testable Hypotheses.\  {\it Lunar and 
Planetary Institute Conference Abstracts} {\bf 35}, 2168. 

\bibitem{} 
Goldberg, S.B., Maimone, M.W. \& Matthies, L. (2002). Stereo Vision and Rover Navigation Software for Planetary Exploration. {\it 2002 IEEE Aerospace Conference Proceedings}, Big Sky, Montana, {\bf 5}, pp. 2025-2036.

\bibitem{} 
Golombek, M., Grant, J., Parker, T., Crisp, J., Squyres, S., Carr, M., 
Haldemann, A., Arvidson, R., Ehlmann, B., Bell, J., Christensen, P., 
Fergason, R., Ruff, S., Cabrol, N., Kirk, R., Johnson, J., Soderblom, L., 
Weitz, C., Malin, M., Rice, J., Anderson, R. \& The Athena Science Team 
(2004). Preliminary Assessment of Mars Exploration Rover Landing Site 
Predictions.\  {\it Lunar and Planetary Institute Conference Abstracts} 
{\bf 35}, 2185. 

\bibitem{} 
Greeley, R., Squyres, S.~W., Arvidson, R.~E., Bartlett, P., Bell, J.~F., 
Blaney, D., Cabrol, N.~A., Farmer, J., Farrand, B., Golombek, M.~P., 
Gorevan, S.~P., Grant, J.~A., Haldemann, A.~F.~C., Herkenhoff, K.~E., 
Johnson, J., Landis, G., Madsen, M.~B., McLennan, S.~M., Moersch, J., Rice, 
J.~W., Richter, L., Ruff, S., Sullivan, R.~J., Thompson, S.~D., Wang, A., 
Weitz, C.~M. \& Whelley, P.\ (2004). Wind-Related Processes Detected by 
the Spirit Rover at Gusev Crater, Mars.\  {\it Science} {\bf 305}, 810-821. 

\bibitem{} 
Gulick, V.~C., Morris, R.~L., Ruzon, M.~A. \& Roush, T.~L.\ (2001). 
Autonomous image analyses during the 1999 Marsokhod rover field test.\  
{\it Journal of Geophysical Research} {\bf 106}, 7745-7764. 

\bibitem{}
Gulick, V.~C., Morris, R.~L., Gazis, ~P.~R., Bishop,  J.~L. \& Alena, R.~L. (2002a).  The Geologist's Field Assistant: Developing an Innovative Science Analysis System for Exploring the Surface of Mars. {\it AGU Fall Meeting Abstracts},  {\bf A364}. 

\bibitem{} 
Gulick, V.~C., Morris, R.~L., Bishop, J., Gazis, P., Alena, R. \& 
Sierhuis, M.\ (2002b). Geologist's Field Assistant: Developing Image and 
Spectral Analyses Algorithms for Remote Science Exploration.\  {\it Lunar 
	and Planetary Institute Conference Abstracts} {\bf 33}, 1961. 

\bibitem{} 
Gulick, V.~C., Morris, R.~L., Gazis, P., Bishop, J.~L., Alena, R., Hart, 
S.~D. \& Horton, A.\ (2003). Automated Rock Identification for Future Mars 
Exploration Missions.\  {\it Lunar and Planetary Institute Conference 
Abstracts} {\bf 34}, 2103. 

\bibitem{} 
Gulick, V.~C., Hart, S.~D., Shi, X. \& Siegel, V.~L.\ (2004). Developing 
an Automated Science Analysis System for Mars Surface Exploration for MSL 
and Beyond.\  {\it Lunar and Planetary Institute Conference Abstracts} {\bf 
35}, 2121. 



\bibitem{}
Haddon, J.F. \& J.F. Boyce. (1990).
Image segmentation by unifying region and boundary information.
{\it IEEE Trans. Pattern Anal.\ Mach.\ Intell.}\ {\bf 12} (10), pp. 929-948.

\bibitem{}
Haralick, R.M.,  Shanmugan, K. \&  Dinstein, I. (1973). Texture features for image classification. {\it IEEE SMC-3} (6), pp. 610-621.


\bibitem{} 
Herkenhoff, K.~E., Squyres, S.~W., Arvidson, R., Bass, D.~S., Bell, J.~F., 
Bertelsen, P., Cabrol, N.~A., Gaddis, L., Hayes, A.~G., Hviid, S.~F., 
Johnson, J.~R., Kinch, K.~M., Madsen, M.~B., Maki, J.~N., McLennan, S.~M., 
McSween, H.~Y., Rice, J.~W., Sims, M., Smith, P.~H., Soderblom, L.~A., 
Spanovich, N., Sullivan, R. \& Wang, A.\ (2004). Textures of the Soils and 
Rocks at Gusev Crater from Spirit's Microscopic Imager.\  {\it Science} 
{\bf 305}, 824-827. 


\bibitem{}
Huntsberger, T., Aghazarian, H., Cheng, Y., Baumgartner, E.T., Tunstel, E., Leger, C., Trebi-Ollennu, A. \& Schenker, P.S. (2002). 
Rover Autonomy for Long Range Navigation and Science Data Acquisition on Planetary Surfaces.
{\it Proceedings of the 2002 IEEE International Conference on Robotics and Automation},  Washington, DC.

\bibitem{}
IGME (Instituto Geol\'ogico y Minero de Espa\~na). (1975). {\it Mapa geologico de Espa\"na E 1:50,000},
Arganda (segunda serie, primera edicion), Hoja numero 583, Memoria explicativa, p. 3-25.

\bibitem{}
Lee, S., Koepp, D., Raimi, S., Arad, A., Robertson, C., {\it et al.}\ (2002, 2004). {\it Spider-man: the Movie, I \& II}, Marvel Comic Books and Sony Pictures. 

\bibitem{} 
Maki, J.~N., Bell, J.~F., Herkenhoff, K.~E., Squyres, S.~W., Kiely, A., 
Klimesh, M., Schwochert, M., Litwin, T., Willson, R., Johnson, A., Maimone, 
M., Baumgartner, E., Collins, A., Wadsworth, M., Elliot, S.~T., Dingizian, 
A., Brown, D., Hagerott, E.~C., Scherr, L., Deen, R., Alexander, D. \& 
Lorre, J.\ (2003). Mars Exploration Rover Engineering Cameras.\  {\it 
Journal of Geophysical Research (Planets)} {\bf 108}, 12-1. 


\bibitem{} 
McEwen, A., Hansen, C., Bridges, N., Delamere, W.~A., Eliason, E., Grant, 
J., Gulick, V., Herkenhoff, K., Keszthelyi, L., Kirk, R., Mellon, M., 
Smith, P., Squyres, S., Thomas, N. \& Weitz, C.\ (2003). MRO's High 
Resolution Imaging Science Experiment (HiRISE): Science Expectations.\  
{\it Sixth International Conference on Mars}, 3217. 


\bibitem{}
McGreevy, M.W. (1992). The presence of field geologists in Mars-like terrain, {\it Presence}, {\bf 1}, no. 4, pp. 375-403. 

\bibitem{}
McGreevy, M.W. (1994). An ethnographic object-oriented analysis of explorer presence in a volcanic terrain environment. {\it NASA Technical Memorandum} \#108823, Ames Research Center, Moffett Field, California. 

\bibitem{}
McGuire, P.~C., Rodriguez Manfredi, J.~A., Sebastian Martinez, E., Gomez Elvira, J., Diaz Martinez, E., Orm\"o, J., Neuffer, K., Giaquinta, A., Camps Martinez, F., Lepinette Malvitte, A., Perez Mercader, J., Ritter, H., Oesker, M., Ontrup, J. \& Walter, J.\ (2004). Cyborg systems as platforms for computer-vision algorithm-development for astrobiology, {\it Proceedings of the III European Workshop on Exo/Astrobiology}, held at the Centro de Astrobiologia, Madrid, http://arxiv.org/abs/cs.CV/0401004, {\it ESA SP-545},  pp. 141-144. 

\bibitem{}
McSween, H.~Y., Arvidson, R.~E., Bell, J.~F., Blaney, D., Cabrol, 
N.~A., Christensen, P.~R., Clark, B.~C., Crisp, J.~A., Crumpler, L.~S., Des 
Marais, D.~J., Farmer, J.~D., Gellert, R., Ghosh, A., Gorevan, S., Graff, 
T., Grant, J., Haskin, L.~A., Herkenhoff, K.~E., Johnson, J.~R., Jolliff, 
B.~L., Klingelhoefer, G., Knudson, A.~T., McLennan, S., Milam, K.~A., 
Moersch, J.~E., Morris, R.~V., Rieder, R., Ruff, S.~W., de Souza, P.~A., 
Squyres, S.~W., W{\" a}nke, H., Wang, A., Wyatt, M.~B., Yen, A., \& 
Zipfel, J.\ (2004). Basaltic Rocks Analyzed by the Spirit Rover in Gusev 
Crater,  {\it Science} {\bf 305}, 842-845. 




\bibitem{}
Moore, J.M. (2004). Blueberry fields for ever. {\it Nature}, {\bf 428}, 711-712.

\bibitem{}
Nesnas, I., Maimone, M., Das, H. (1999). Autonomous Vision-Based Manipulation from a Rover Platform, {\it Proceedings of the CIRA Conference}, Monterey, California.

\bibitem{}
Olson, C.F., Matthies, L.H., Schoppers, M. \& Maimone, M.W.  (2003). Rover Navigation Using Stereo Ego-motion.  {\it Robotics and Autonomous Systems}, {\bf 43}(4), 215-229. 

\bibitem{}
Pedersen, L., (2001). Autonomous characterization of unknown environments. {\it 2001 IEEE International Conference on Robotics and Automation}, {\bf 1}, 277-284. 

\bibitem{}
Rae, R.,  Fislage, M.  \& Ritter, H. (1999).
Visuelle Aufmerksamkeitssteuerung zur Unterst\"utzung gestikbasierter 
Mensch-Maschine Interaktion,
{\it KI - K\"unstliche Intelligenz, Themenheft Aktive Sehsysteme}, Vol. 01, March issue, Hrsg.  B\"arbel Mertsching, pp. 18-24.


\bibitem{}
Sebe, N.,  Tian, Q.,  Loupias, E., Lew, M. \& Huang, T.S. (2003). Evaluation of Salient Points Techniques, {\it Image and Vision Computing, Special Issue on Machine Vision} {\bf 21}, 1087-1095.


\bibitem{} 
Squyres, S.~W., Arvidson, R.~E., Baumgartner, E.~T., Bell, J.~F., 
Christensen, P.~R., Gorevan, S., Herkenhoff, K.~E., Klingelh{\" o}fer, G., 
Madsen, M.~B., Morris, R.~V., Rieder, R. \& Romero, R.~A.\ (2003). Athena 
Mars rover science investigation.\  {\it Journal of Geophysical Research 
(Planets)} {\bf 108}, 3-1. 






\bibitem{} 
Squyres, S.\ (2004). Initial Results from the MER Athena Science 
Investigation at Gusev Crater and Meridiani Planum.\  {\it AGU Spring 
Meeting Abstracts}, A1. 


\bibitem{}
Squyres, S. W. \& Athena Science Team (2004). Science Results From The MER Athena Science Investigation At Gusev Crater And Meridiani Planum. Abstract to the {\it Klein Lecture} at Astrobiology Science Conference, {\it International Journal of Astrobiology}, {\bf 3}, Supplement, p. 5.

\bibitem{} 
Squyres, S.~W., Arvidson, R.~E., Bell, J.~F., Br{\" u}ckner, J., Cabrol, 
N.~A., Calvin, W., Carr, M.~H., Christensen, P.~R., Clark, B.~C., Crumpler, 
L., Des Marais, D.~J., d'Uston, C., Economou, T., Farmer, J., Farrand, W., 
Folkner, W., Golombek, M., Gorevan, S., Grant, J.~A., Greeley, R., 
Grotzinger, J., Haskin, L., Herkenhoff, K.~E., Hviid, S., Johnson, J., 
Klingelh{\" o}fer, G., Knoll, A., Landis, G., Lemmon, M., Li, R., Madsen, 
M.~B., Malin, M.~C., McLennan, S.~M., McSween, H.~Y., Ming, D.~W., Moersch, 
J., Morris, R.~V., Parker, T., Rice, J.~W., Richter, L., Rieder, R., Sims, 
M., Smith, M., Smith, P., Soderblom, L.~A., Sullivan, R., W{\" a}nke, H., 
Wdowiak, T., Wolff, M. \& Yen, A.\ (2004). The Spirit Rover's Athena 
Science Investigation at Gusev Crater, Mars.\  {\it Science} {\bf 305}, 
794-800. 


\bibitem{}
Storrie-Lombardi, M.C., Grigolini, P., Galatolo, S., Tinetti, G., Ignaccolo, M., Allegrini, P. \& Corsetti, F.A. (2002). Advanced Techniques in Complexity Analysis for the Detection of Biosignatures in Ancient and Modern Stromatolites. {\it Astrobiology}, {\bf 2}(4), pp. 630-631.

\bibitem{}
Vandapel, N., Chatila, R., Moorehead, S., Lacroix, S., Apostolopoulos, D. 
\& W.L. Whittaker, W.L. (2000).
Evaluation of Computer Vision Algorithms for Autonomous Navigation in Polar Terrains. {\it International Conference on Intelligent Autonomous Systems}.


\bibitem{}
Volpe, R. (2003). Rover Functional Autonomy Development for the Mars Mobile Science Laboratory. {\it Proceedings of the 2003 IEEE Aerospace Conference}, Big Sky, Montana.

\bibitem{}
Wagner, M.D. (2000).
Experimenter's Notebook: Robotic Search for Antarctic Meteorites 2000 Expedition. Technical report CMU-RI-TR-00-13, Robotics Institute, Carnegie Mellon University. 

\bibitem{}
Wettergreen, D., Bapna, D., Maimone, M. \& Thomas, H. (1999). Developing Robotic Exploration of the Atacama Desert. {\it Robotics and Autonomous Systems}, {\bf 26}(2-3), pp. 127-148.

\bibitem{}
Whittaker, W.L., Bapna, D., Maimone, M. \& Rollins, E. (1997).
Atacama Desert Trek: A Planetary Analog Field Experiment. {\it Proceedings of the Fourth International Symposium on Artificial Intelligence, Robotics and Automation for Space (i-SAIRAS'97)}, Tokyo, Japan, 

\bibitem{}
Wrede, S., Hanheide, M., Bauckhage, C. \& Sagerer, G. (2004). An Active Memory as a Model for Information Fusion. {\it Proceedings of the 7th International FUSION Conference}. 

\end{thebibliography}
\end{document}